\documentclass{egpubl}
\usepackage{eg2018}

\ConferencePaper
 \electronicVersion 

\ifpdf \usepackage[pdftex]{graphicx} \pdfcompresslevel=9
\else \usepackage[dvips]{graphicx} \fi

\PrintedOrElectronic

\usepackage{t1enc,dfadobe}

\usepackage{egweblnk}
\usepackage{cite}

\usepackage{amsmath}
\usepackage{amssymb}
\usepackage{multirow}

\newcommand\la[1]{\multicolumn{1}{l}{#1}}
\newcommand\mb[1]{\mathbf{#1}}

\newcommand\mi[1]{\mathit{#1}}

\usepackage{booktabs}
\usepackage{subcaption}
\usepackage{tikz}
\usetikzlibrary{fit}
\usetikzlibrary{shapes,arrows}

\makeatletter\def\floatc@ruled#1#2{#2\par}\makeatother

\newcommand{\tc}[1]{{#1}}

\title[ExpandNet]%
      {ExpandNet: A Deep Convolutional Neural Network for High Dynamic Range Expansion from Low Dynamic Range Content}

\author[Marnerides et al.]
{\parbox{\textwidth}{\centering D. Marnerides$^{1,2}$,
T. Bashford-Rogers$^{3}$,
J. Hatchett$^{2}$
and K. Debattista$^{2}$
        }
        \\
{\parbox{\textwidth}{\centering $^1$Warwick Centre for Predictive Modelling (WCPM), University of Warwick, UK\\
         $^2$ WMG, University of Warwick, UK\\
         $^3$ Department of Computer Science and Creative Technologies, University of the West of England, UK
       }
}
}

\begin{document}


\maketitle
\begin{abstract}
    High dynamic range (HDR) imaging provides the capability of handling real world
    lighting as opposed to the traditional low dynamic range (LDR) which struggles
    to accurately represent images with higher dynamic range. However, most imaging
    content is still available only in LDR. This paper presents a method for
    generating HDR content from LDR content based on deep Convolutional Neural
    Networks (CNNs) termed ExpandNet. ExpandNet accepts LDR images as input and
    generates images with an expanded range in an end-to-end fashion. The model
    attempts to reconstruct missing information that was lost from the original
    signal due to quantization, clipping, tone mapping or gamma correction. The
    added information is reconstructed from learned features, as the network is
    trained in a supervised fashion using a dataset of HDR images. The approach is
    fully automatic and data driven; it does not require any heuristics or human
    expertise. ExpandNet uses a multiscale architecture which avoids the use of
    upsampling layers to improve image quality. The method performs well compared to
    expansion/inverse tone mapping operators quantitatively on multiple
    metrics, even for badly exposed inputs.


\end{abstract}

\section{\textbf{Introduction}}

High dynamic range (HDR) imaging provides the capability to capture, manipulate and display real-world lighting, unlike traditional, low dynamic
range (LDR) imaging. HDR has found many applications in photography, physically-based rendering, gaming, films, medical and industrial imaging and
recent displays support HDR content~\cite{eetzen2004hdr, marchessoux2016medicalhdr}. While HDR imaging has seen many advances, LDR remains the status
quo, and the majority of both current and legacy content is predominantly LDR. In order to gain an improved viewing experience~\cite{akyuz06}, or to
use this content in future HDR pipelines, LDR content needs to be converted to HDR.

A number of methods which can retarget LDR to HDR content have been
presented~\cite{banterle2011hdrbook}. These methods make it possible to utilise
and manipulate the vast amounts of LDR content within HDR pipelines and
visualise them on HDR displays. However, such methods are primarily
model-driven, use various parameters which make them difficult to use by
non-experts, and are not suitable for all types of content.

Recent machine learning advances for applications in image processing provide
data driven solutions for imaging problems, bypassing reliance on human
expertise and heuristics. CNNs are the current de-facto approach used for many
imaging tasks, due to their high learning capacity as well as their
architectural qualities which make them highly suitable for image
processing~\cite{schmidhuber2014deep}. The networks allow for abstract
representations to be acquired directly from data, surpassing simplistic
pixelwise processing. This acquisition of abstractness is especially strong
when the networks are of sufficient depth~\cite{he2015residual}. This paper
presents a method for HDR expansion based on deep Convolutional Neural Networks
(CNNs).

In this work, a novel multiscale CNN architecture, called ExpandNet, is
presented. On a local scale, one branch of the network learns how to maintain
and expand high frequency detail, while a dilation branch learns information
on larger pixel neighbourhoods. A final third branch provides overall
information by learning the global context of the input. The architecture is
designed to avoid upsampling of downsampled features, in an attempt to reduce
blocking and/or haloing artefacts that may arise from more straightforward
approaches, for example autoencoder architectures~\cite{bengio2009deep}.
Results demonstrate an improvement in quality over all other previous
approaches that were tested, including some other CNN architectures.

\noindent In summary, the primary contributions of this work are:
\begin{itemize}
  \item A fully automatic, end-to-end, parameter free method for the expansion
      of LDR content based on a novel CNN architecture which improves image
      quality for HDR expansion.
  \item Results which are competitive with the other approaches tested, including other CNN architectures applied to single exposure LDR to HDR.
  \item Data augmentation for limited HDR content via different exposure and position selection to obtain more LDR-HDR training pairs.
  \item A comprehensive quantitative comparison of LDR to HDR expansion methods.
\end{itemize}

\section{\textbf{Related Work}}

A number of methods to expand LDR to HDR have been presented in the literature. Furthermore, deep learning methods have been used for similar
problems in the past. The following subsections discuss these topics.

\subsection{\textbf{LDR to HDR}}
Expansion operators (EOs), also known as inverse or reverse tone mapping operators, attempt to generate HDR content from LDR content. EOs can
generally be expressed as:

\begin{equation}
L_e = f(L_d), \text{ where } f:  [0,255] \to \mathbb{R}^{+}
\label{eqn:eo}
\end{equation}

\noindent where $L_e$ corresponds to the expanded HDR content, $L_d$ to the LDR input and $f(\cdot)$ is the EO. In this context $f(\cdot)$ could be
considered as an ill-posed function. However, a variety of methods have emerged that attempt to tackle this issue. The majority of EOs can be broadly
divided into two categories: global and local methods~\cite{banterle2011hdrbook}.

The global methods use a straightforward function to expand the content equally
across all pixels. One of the first of such methods was the technique presented
by Landis~\cite{Landis02} which expands content based on power functions. A
straightforward method that uses a linear transformation combined with gamma
correction was presented by Aky\"{u}z et al.~\cite{akyuz06} and evaluated using
a subjective experiment. Masia et al.~\cite{masia09,masia2017dynamic} also
presented a global method which expands the content based on image attributes
defined by an image key.

Local methods typically expand LDR content to HDR through the use of an
analytical function combined with an expand map. The inverse tone mapping
method~\cite{banterle06inverse} initially expands the content using an inverted
photographic tone reproduction tone mapper~\cite{reinhard2002tmo}, although this could be applied to other tone
mappers that are invertible. An expand map is generated by selecting a constellation of bright
points and expanding them via density estimation. This is subsequently used in
conjunction with the inverse tone mapping equation to map LDR values to HDR
values to avoid quantization errors that would arise via inverse tone mapping
only. Rempel et al.~\cite{rempel06ldr2hdr} also used an expand map, however
this was computed through the use of a Gaussian filter in conjunction with an
edge-stopping function to maintain contrast. Kovaleski and Oliviera
\cite{Kovaleski+2014} extended the work of Rempel et al.\ via the use of a cross
bilateral filter. Subsequently, Huo et al.~\cite{Huo+2014} further extended
this work to remove the thresholding used by Kovaleski and Oliviera.

Other methods include inpainting as used by Wang et al.~\cite{Wang+2007} which is partially user-based, and classification based methods such as by
Meylan et al.~\cite{Meylan+2006} and Didyk et al.~\cite{Didyk+2008}, which operate on different parts of the image by classifying these parts
accordingly.

Banterle et al.~\cite{banterle2009high} provide a broader view of these
methods. With most of the above, the added information is derived from
heuristics that may produce sufficient results for well behaved inputs, but are
not data driven. Most importantly, most existing EOs find it difficult to
handle under/over-exposed LDR content.

\subsection{\textbf{Deep Learning for Image Processing}}

Deep learning has been extensively used for image processing problems recently.
In image-to-image translation~\cite{isola2016pix2pix} the authors present a method based on Generative
Adversarial Networks~\cite{goodfellow2015gan} and the
U-Net~\cite{ronneberger2015unet} architecture that transforms images from one
domain to another (e.g.~maps to satellite). Many approaches have also been
developed for other kinds of ill-posed or inverse problems, including image
super-resolution and
upsampling~\cite{dong2016sr,kim2015drsr,yamanaka2017fastsr} as well as
inpainting/hallucination of missing information~\cite{iizuka2017completion}.
Automatic colorization~\cite{iizuka2016colornet} converts grey scale to color images using a CNN which
uses two routes of computation, fusing local and global context for improved
image quality.

In visualization, graphics and HDR imaging, neural networks have been used for
predicting sky illumination for
rendering~\cite{satilmis2015machine,hold2016illum}, denoising Monte Carlo
renderings~\cite{kalantari2015machine,chaitanya2017interactive,bako2017kernel},
predicting HDR environment maps~\cite{zhang2017learninghdr}, reducing artefacts
such as ghosting when fusing multiple LDR exposures to create HDR
content~\cite{kalantari2017hdr} and for tone mapping~\cite{hou2017tonemap}.

\tc{Concurrently to this work, two other deep learning approaches that expand
LDR content to HDR have been developed. Eilertsen et
al.~\cite{eilertsen2017cnn}, use a U-Net like architecture to predict values for
saturated areas of badly exposed content, whereas non-saturated areas are
linearised by applying an inverse camera response curve. Endo et
al.~\cite{endo2017drtmo} use a modified U-Net architecture that predicts
multiple exposures from a single exposure which are then used to generate an HDR image using
standard merging algorithms.}

\tc{The first method does not suffer greatly from artefacts produced from
upsampling that are common with U-Net and similar
architectures~\cite{odena2016deconvolution} since only areas of badly exposed
content are expanded by the network. In the latter, the authors mention the
appearance of tiling artefacts in some cases. There are other examples in
literature when fully converged U-Net like networks exhibit artefacts, for
example in image-to-image translation tasks~\cite{pix2pixsuppl}, or semantic
segmentation~\cite{zhang2017suppl}. Our approach differs from these methods
methods as it presents a dedicated architecture for and end-to-end image
expansion, without using upsampling. } 

\begin{figure*}[htb]
    \centering
    \includegraphics[width=0.9\linewidth]{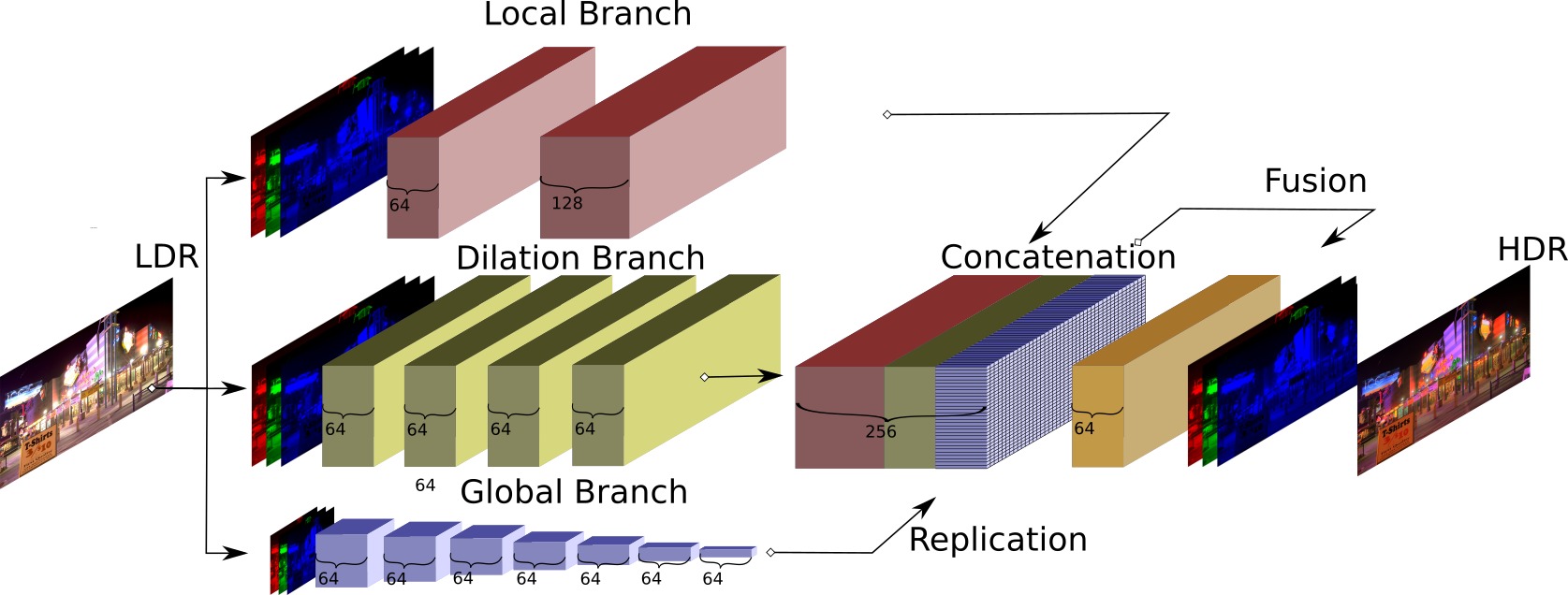}
    \caption{ExpandNet architecture. The LDR input is propagated through the
    the local and dilation branches, while a resized input ($256\times256$) is
    propagated through the global branch. The output of the global branch is
    superposed over each pixel of the outputs of the other two branches. The
    resulting features are fused using $1 \times 1$ convolutions to form the
    last feature layer which then gives an RGB HDR
    prediction.}\label{fig:network}
\end{figure*}

\section{\textbf{ExpandNet}}

This section describes the ExpandNet architecture in detail. The network is
designed to tackle the problem directly via a novel three branch architecture.
Figure~\ref{fig:network} presents an overview of the architecture. The three branches of computation are a local, a dilation and a global one.
Each branch is itself a CNN that accepts an RGB LDR image as input. Each one of the
three branches is responsible for a particular aspect, with the local branch
handling local detail, the dilation branch for medium level detail, and a
global branch accounting for higher level image-wide features.

The local and dilation branches avoid any use of downsampling and upsampling, which is a common approach in the design of CNNs, and the global branch
only downsamples. In image processing CNNs it is common to downsample the width and height of the input image, while expanding the channel dimension.
This forms a set of more abstract features after a few layers of downsampling. The features are then upsampled to the original dimensions, for
example in autoencoders. \tc{As also mentioned in the previous section, it is} argued~\cite{odena2016deconvolution} that upsampling, especially the frequently used deconvolutional
layers~\cite{shi2016deconv}, cause checkerboard artefacts. Furthermore, upsampling may cause unwanted information bleeding in areas where context is
missing, for example large over-exposed areas. Figure~\ref{fig:hallucinations_low} and Figure~\ref{fig:hallucinations_grotto} (b) and (c), discussed further in Section~\ref{sec:results},
provide examples where such artefacts can arise in upsampling networks, seen as blocking in (b) due to deconvolutions, and banding in (c) due to
nearest-neighbour upsampling. ExpandNet avoids the use of upsampling layers to reduce such artefacts and improves the quality of the predicted HDR
images.

The outputs of the three branches are fused and further processed by a small
final convolutional layer that produces the predicted HDR image. The input LDR
and the predicted HDR are both in the $[0, 1]$ range.

The following subsection briefly introduces CNNs, followed by a detailed
overview of the three branches of the ExpandNet architecture, including design
characteristics for feature fusion, activation functions and the loss function
used for optimization.

\begin{figure*}[htb]
    \centering
    \includegraphics[width=0.9\linewidth]{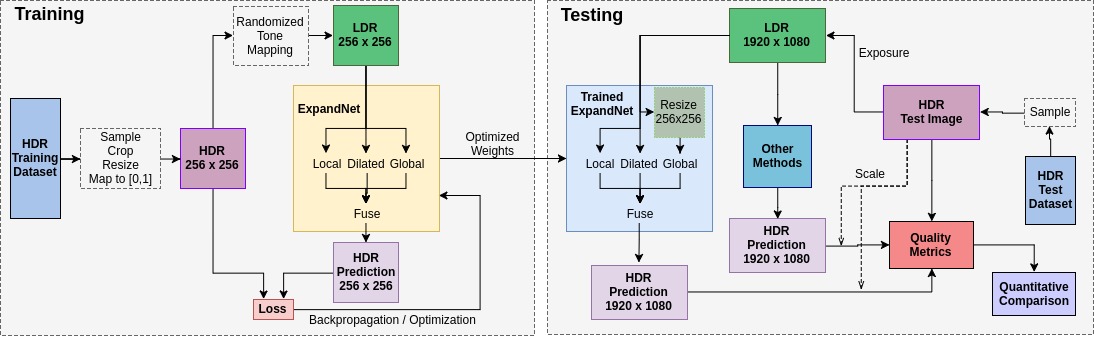}
    \caption{General overview of the workflow. (left) The training dataset is
    sampled and preprocessed on-the-fly to form $256\times256$ resolution
    input-output pairs, which are then used to optimize the network weights.
    (right) For testing, the images are full-HD ($1,920 \times 1,080$). The
    luminance of the predictions of all methods is scaled either to match the
    original HDR image (scene-referred) or that of a 1,000 $cd/m^2$ display
    (display-referred).}\label{fig:workflow}
\end{figure*}

\subsection{\textbf{Convolutional Neural Networks}}

A feed-forward neural network (NN) is a function composed of multiple layers of
non-linear transformations. Given an input vector $\mb{x}$, a network of $M$
layers (with no skip connections) can be expressed as follows:
\begin{equation}
    \mi{f}_{NN}(\mb{x}) = (l_M \circ l_{M-1} \circ \dots \circ l_2 \circ l_1)(\mb{x})
\end{equation}
where $l_i$ is the $i^{th}$ hidden layer of the network and $\circ$ is the
composition operator. Each layer accepts the output of the previous layer,
$\mb{o}_{i-1}$, and applies a linear map followed by a non-linear
transformation:
\begin{equation}
    \mb{o}_i = l_i(\mb{o}_{i-1}) = \alpha (W_i \mb{o}_{i-1})
\end{equation}
where $W_i$ is a matrix of learnable parameters (weights), $\mb{o}_N$ is the
network output and $\mb{o}_0 = \mb{x}$. $\alpha(z)$ is a non-linear (scalar)
activation function, applied to each value of the resulting vector
independently. A learnable bias term exists in the linear map as well, but is
folded in $W_i$ (and $\mb{x}$) for ease of notation.

A convolutional layer, $c_i$, uses sparse parameter matrices with repeated
values. The sparsity and repetition structure is such, so that the linear
product can be expressed as a convolution, $\ast$, between a learnable parameter filter
$\tilde{w}$ and the input to the layer.
\begin{equation}
    \mb{c}_i (\mb{o}_{i-1}) = \alpha (\tilde{w}_{i} \ast \mb{o}_{i-1})
\end{equation}

This formulation is analogous for higher dimensions. In the scope of this work,
images are three dimensional objects
(width $\times$ height $\times$ channels / features), thus the parameter matrices
become four-dimensional tensors. For image processing CNNs, the convolutions
are usually only in the width and height dimensions, while the third dimension
is fully connected (dense tensor dimension).

The convolutional architecture is extremely suitable for images since it exploits spatial correlations and symmetries, and dramatically reduces the
number of learnable parameters compared to fully connected networks. It also allows for efficient implementations on GPUs as well as more stable
training of deeper models~\cite{schmidhuber2014deep}.

\subsection{\textbf{Branches}}

The three branches play different roles in expanding the dynamic range of the
input LDR. The global branch seeks to reduce the dimensionality of the input
and capture abstract features. It has a sufficiently large receptive field that
covers the whole image. It accepts the entire LDR image as input, re-sized to
$256\times256$, and eventually downsamples it to $1\times1$ over a total of
seven layers. Each layer has $64$ feature maps and uses stride $2$ convolutions
which consecutively downsample the spatial dimensions by a factor of $2$. All
the global branch layers use a convolutional kernel of size $3 \times 3$, with
padding $1$ except the last layer which uses a $4 \times 4$ kernel with no
padding, essentially densely connecting the previous layer, which consists of
$4\times4$ features, with the last layer, creating a vector of $1\times1$
features.

The other two branches provide localized processing without downsampling that
captures higher frequencies and neighbouring features. The local branch has a
receptive field of $5\times5$ pixels and consists of two layers with $3\times3$
convolutions of stride $1$ and padding $1$, with $64$ and $128$ feature maps
respectively. The small receptive field of the local branch provides learning
at the pixel level, preserving high frequency detail.

The dilation branch has a wider receptive field of $17\times17$ pixels and uses
dilated convolutions~\cite{yu2015dilated} of dilation size $2$, kernel
$3\times3$, stride $1$, and padding $2$. Dilated convolutions are large, sparse
convolutional kernels, used to quickly increase the receptive field of CNNs. A
total of four dilation layers are used each with $64$ features. With an
increased receptive field, the dilation network captures local features with
medium range frequencies otherwise missed by the other two branches whose focus
is on the two extremes of the frequency spectrum.

\begin{figure*}[htb]
    \centering
    \begin{subfigure}[t]{0.3\linewidth}
        \centering
        \includegraphics[width=1.0\linewidth]{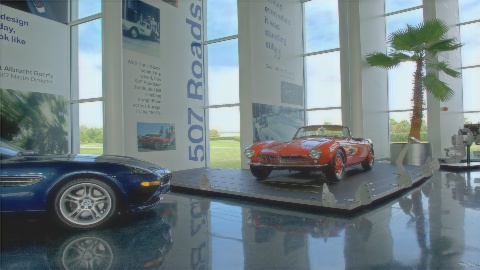}
        \caption{Local + Dilated + Global}
    \end{subfigure}
    \begin{subfigure}[t]{0.3\linewidth}
        \centering
        \includegraphics[width=1.0\linewidth]{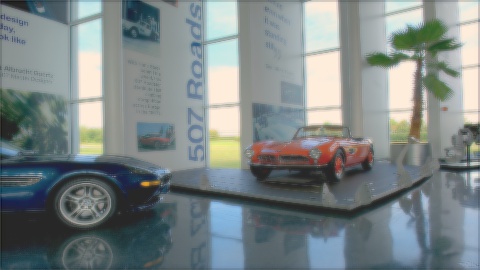}
        \caption{Local + Global}\label{fig:branches:LG}
    \end{subfigure}
    \begin{subfigure}[t]{0.3\linewidth}
        \centering
        \includegraphics[width=1.0\linewidth]{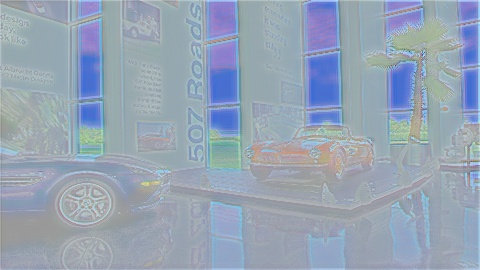}
        \caption{Dilated + Global}\label{fig:branches:DG}
    \end{subfigure}\\
    \begin{subfigure}[t]{0.3\linewidth}
        \centering
        \includegraphics[width=1.0\linewidth]{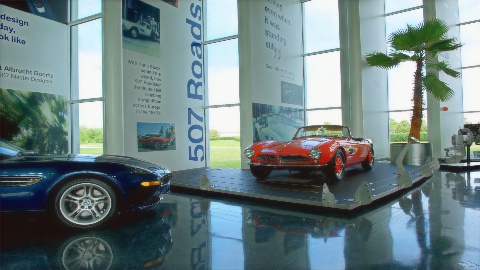}
        \caption{Local + Dilated}
    \end{subfigure}
    \begin{subfigure}[t]{0.3\linewidth}
        \centering
        \includegraphics[width=1.0\linewidth]{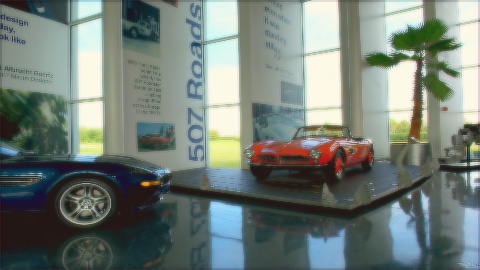}
        \caption{Local}
    \end{subfigure}
    \begin{subfigure}[t]{0.3\linewidth}
        \centering
        \includegraphics[width=1.0\linewidth]{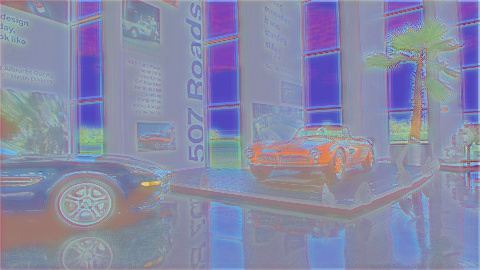}
        \caption{Dilated}
    \end{subfigure}
    \caption{Illustration of the contribution of each of the three branches of
    ExpandNet. These images were obtained by masking one or more branches with
    zero inputs. The bottom row is produced with the global branch masked. This
    causes the overall appearance of the images to be darker and sharper, since
    there are low frequencies missing. The middle column masks the dilation
    branch, resulting in sharp high-frequency images. The right column masks
    the local branch which causes most of the fine details to be
    lost.}\label{fig:branches}
\end{figure*}

The effects of each individual branch are presented in
Figure~\ref{fig:branches}. Masking the input to an individual branch causes the
output appearance to change, depending on which branch was masked, highlighting
its role. The local branch produces high frequency features,
while the dilation branch adds medium range frequencies. The global branch
changes the overall appearance of the output by adding low frequencies and
adjusting the overall sharpness of the image. \tc{Results, shown in Section~\ref{sec:results:further}, further 
help to illustrate the advantages posed by the three distinct branches.}

\subsection{\textbf{Fusion}}

The outputs of the three branches are merged in a manner similar to the fusion
layer by Iizuka et al.~\cite{iizuka2016colornet}. The local and dilation outputs, which have
the same height and width as the input, are concatenated along the feature map
dimension. The output of the global network is a vector of $64$ features which
is replicated along the width and height dimensions to match the dimensions of
the other two outputs. The replication superposes the vector over each pixel of
the predictions of the other two branches. It is then concatenated with the
rest of the outputs along the feature map dimension resulting in a total of
$256$ features. The concatenation is followed by a convolution of kernel size
$1\times 1$ which fuses the global feature vector with each individual pixel of
the local and dilated features, thus combining context from multiple scales.
The output of the fusion layer is further processed by a final convolutional
layer with $3\times3$ kernels, stride $1$ and padding $1$.

\subsection{\textbf{Activations}}

All the layers, besides the output layer, use the Scaled Exponential Linear
Unit (SELU) activation function~\cite{klambauer2017selu}, a variation of the
Exponential Linear Unit (ELU).
\begin{equation}
    \text{SELU}(z) = \beta \begin{cases}
        z &\text{if $z > 0$}\\
        \alpha e^z  - \alpha &\text{if $z \leq 0$}
        \end{cases}
\end{equation}
where $\beta \approx 1.05070$ and $\alpha \approx 1.67326$. SELU was recently introduced for the creation of self normalizing neural networks and it
ensures that the distributions of the activations at each layer have a mean of zero and unit variance. It provides a solution to the internal
covariate shift problem during training at a lower memory cost compared to the frequently used batch normalization
technique~\cite{ioffe2015batchnorm}. The SELU unit also preserves all the properties of the ELU, which in its turn improves on the Rectified Linear
Unit (ReLU). ReLUs alleviate the vanishing/exploding gradient problem~\cite{krizhevsky2017alexnet} that was frequent with the traditional Sigmoid
activations (when stacked), while ELUs improve the sparse activation problem of the ReLUs by providing negative activation values.

The final layer of the network uses a Sigmoid activation,
\begin{equation}
    \sigma(z) = \frac{1}{1 + e^{-z}}
\end{equation}
which maps the output to the $[0, 1]$ range.

\begin{table}[!t]
    \caption{Parameters used for tone mapping. All images are followed by a gamma
    correction curve with\ $\gamma \in [1.8,2.2]$. Values given within ranges are
    sampled from a uniform distribution.}\label{table:tmo_param}
    \centering
    \begin{tabular}{ll}
    \toprule
    \bfseries TMO & \bfseries Parameters\\\midrule
    \la{\multirow{3}{*}{Photoreceptor}} & \la{Intensity: $[-1.0,1.0]$}\\
                                   & \la{Light adaptation: $[0.8,1.0]$}\\
                                   & \la{Color adaptation: $[0.0,0.2]$}\\\midrule
    \la{\multirow{1}{*}{ALM}}    & \la{Saturation:  $1.0$, Bias: $[0.7,0.9]$}\\\midrule
    \la{\multirow{1}{*}{Display Adaptive}}  & \la{Saturation:  $1.0$, Scale: $[0.65,0.85]$}\\\midrule
    \la{\multirow{2}{*}{Bilateral}}   & \la{Saturation:  $1.0$, Contrast: $[3,5]$}\\
                                   & \la{$\sigma_{\text{space}}: 8$, $\sigma_{\text{color}}: 4$}\\\midrule
    \la{\multirow{1}{*}{Exposure}} & \la{Percentile: $[0,15]$ to $[85,100]$}\\\bottomrule
    \end{tabular}
\end{table}

\subsection{\textbf{Loss function}}

The Loss function, $\mathcal{L}$, used for optimizing the network
is the $L_1$ distance between the predicted image, $\tilde{I}$, and real HDR image,
$I$, from the dataset. The $L_1$ distance is chosen for this
problem since the more frequently used $L_2$ distance was found to cause blurry
results for images~\cite{mathieu2016mse}. An additional cosine similarity term
is added to ensure color correctness of the RGB vectors of each pixel.

\begin{equation}
    \mathcal{L}_{i} =  \lVert \tilde{I}_{i} - I_{i}
    \rVert_{1} + \lambda  \left( 1 -
    \frac{1}{K}\sum_{j=i}^{K}
    \frac{\tilde{I}_{i}^{j} \cdot I_{i}^{j}}{\lVert
    \tilde{I}_{i}^{j}\rVert_{2}\lVert I_{i}^{j}\rVert_{2}}  \right)
    \label{eq:Loss}
\end{equation}
where $\mathcal{L}_{i}$ is the loss contribution of the $i^{\text{th}}$ image
of the dataset, $\lambda$ is a constant factor that adjusts the contribution of
the cosine similarity term, $I_{i}^{j}$ is the $j^{\text{th}}$ RGB pixel vector
of image $I_i$ and $K$ is the total number of pixels of the image.

Cosine similarity measures how close two vectors are by comparing the angle
between them, not taking magnitude into account. For the context of this work,
it ensures that each pixel points in the same direction of the three
dimensional RGB space. It provides improved color stability, especially for low
luminance values, which are frequent in HDR images, since slight variations in any of
the RGB components of these low values do not contribute much to the $L_1$
loss, but they may however cause noticeable color shifts. 

\subsection{\textbf{Training}}

\begin{table}[tbp]
    \caption{Average values of the four metrics for all methods for
    scene-referred scaling. Bold values indicate the best
    value.}\label{table:resultsscene}
    \centering
    \begin{tabular}{lcccc}
        \toprule
        \la{{Method}}&SSIM&MS-SSIM&PSNR&HDR-VDP-2.2\\\midrule
        \multicolumn{5}{c}{\textit{optimal}} \\\midrule
        \la{LAN}&$0.72$&$0.78$&$22.21$&$39.01$\\
        \la{AKY}&$0.72$&$0.78$&$22.70$&$39.11$\\
        \la{MAS}&$\mathbf{0.75}$&$\mathbf{0.80}$&$23.29$&$38.98$\\
        \la{BNT}&$0.70$&$0.73$&$19.56$&$37.63$\\
        \la{KOV}&$0.74$&$\mathbf{0.80}$&$25.03$&$38.39$\\
        \la{HUO}&$0.74$&$0.78$&$19.71$&$38.04$\\
        \la{REM}&$0.68$&$0.64$&$15.68$&$33.61$\\
        \la{COL}&$0.58$&$0.69$&$23.21$&$31.23$\\
        \la{UNT}&$0.68$&$0.71$&$20.52$&$34.88$\\
        \la{EIL}&$0.72$&$0.78$&$22.90$&$39.06$\\
        \la{EXP}&$0.74$&$0.79$&$\mathbf{25.54}$&$\mathbf{39.27}$\\\midrule
        \multicolumn{5}{c}{\textit{culling}} \\\midrule
        \la{LAN}&$0.72$&$0.64$&$17.15$&$30.47$\\
        \la{AKY}&$0.72$&$0.64$&$17.08$&$30.75$\\
        \la{MAS}&$0.72$&$0.63$&$16.87$&$30.59$\\
        \la{BNT}&$0.74$&$0.66$&$18.91$&$32.03$\\
        \la{KOV}&$0.75$&$0.68$&$18.60$&$31.92$\\
        \la{HUO}&$0.75$&$0.64$&$16.27$&$29.95$\\
        \la{REM}&$0.63$&$0.49$&$13.55$&$27.34$\\
        \la{COL}&$0.63$&$0.69$&$22.08$&$29.74$\\
        \la{UNT}&$0.77$&$0.70$&$19.66$&$34.65$\\
        \la{EIL}&$0.52$&$0.53$&$17.92$&$28.14$\\
        \la{EXP}&$\mathbf{0.81}$&$\mathbf{0.79}$&$\mathbf{22.58}$&$\mathbf{35.04}$\\
        \bottomrule
    \end{tabular}
\end{table}
\begin{table}[tbp]
    \caption{Average values of the four metrics for all methods for
    display-referred scaling. Bold values indicate the best
    value.}\label{table:resultsdisplay}
    \centering
    \begin{tabular}{lcccc}
        \toprule
        \la{{Method}}&SSIM&MS-SSIM&PSNR&HDR-VDP-2.2\\\midrule
        \multicolumn{5}{c}{\textit{optimal}} \\\midrule
        \la{LAN}&$0.76$&$0.80$&$19.89$&$\mathbf{41.01}$\\
        \la{AKY}&$0.76$&$0.80$&$20.37$&$40.89$\\
        \la{MAS}&$0.79$&$ 0.82 $&$21.03$&$40.83$\\
        \la{BNT}&$0.74$&$0.75$&$17.22$&$39.99$\\
        \la{KOV}&$\mathbf{0.80}$&$ \mathbf{0.83}$&$23.01$&$40.00$\\
        \la{HUO}&$0.77$&$0.77$&$17.83$&$38.58$\\
        \la{REM}&$0.66$&$0.59$&$14.60$&$33.74$\\
        \la{COL}&$0.63$&$0.71$&$21.00$&$31.41$\\
        \la{UNT}&$0.72$&$0.73$&$18.23$&$35.68$\\
        \la{EIL}&$0.77$&$0.80$&$20.66$&$\mathbf{41.01}$\\
        \la{EXP}&$0.79$&$ 0.82$&$\mathbf{23.43}$&$40.81$\\\midrule
        \multicolumn{5}{c}{\textit{culling}} \\\midrule
        \la{LAN}&$0.31$&$0.17$&$9.12$&$18.01$\\
        \la{AKY}&$0.74$&$0.66$&$15.00$&$31.39$\\
        \la{MAS}&$0.73$&$0.64$&$14.77$&$31.11$\\
        \la{BNT}&$0.36$&$0.27$&$9.61$&$24.51$\\
        \la{KOV}&$0.77$&$0.69$&$16.54$&$31.78$\\
        \la{HUO}&$0.74$&$0.64$&$14.85$&$30.57$\\
        \la{REM}&$0.59$&$0.46$&$12.81$&$27.96$\\
        \la{COL}&$0.66$&$0.70$&$\mathbf{19.99}$&$30.26$\\
        \la{UNT}&$0.78$&$0.69$&$17.02$&$35.27$\\
        \la{EIL}&$0.54$&$0.55$&$15.96$&$27.58$\\
        \la{EXP}&$\mathbf{0.83}$&$\mathbf{0.79}$&$19.93$&$\mathbf{36.21}$\\
        \bottomrule
    \end{tabular}
\end{table}

\section{Training and Implementation}

This section presents the implementation details used for ExpandNet, including
the dataset used and how it was augmented, and implementation and optimization
details. Results are presented in Section~\ref{sec:results}.
Figure~\ref{fig:workflow} gives an overview of the training and testing
methodology employed.

\begin{figure}[t]
    \centering
    \begin{minipage}{0.48\linewidth}
        \centering
        \includegraphics[width=1.0\linewidth]{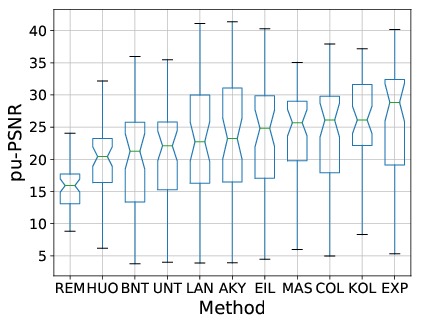}
    \end{minipage}
    \begin{minipage}{0.48\linewidth}
        \centering
        \includegraphics[width=1.0\linewidth]{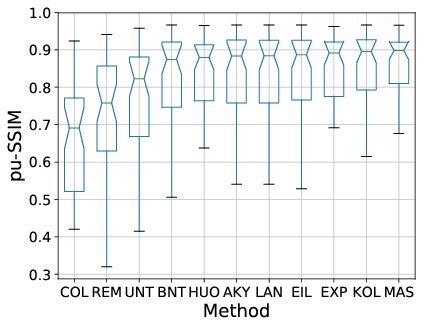}
    \end{minipage}\\
    \begin{minipage}{0.48\linewidth}
        \centering
        \includegraphics[width=1.0\linewidth]{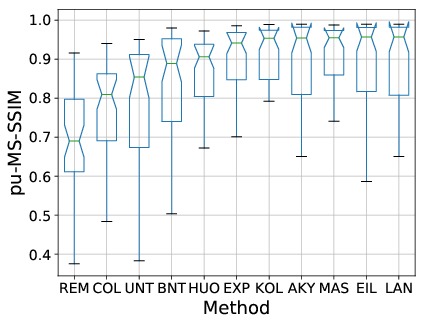}
    \end{minipage}
    \begin{minipage}{0.48\linewidth}
        \centering
        \includegraphics[width=1.0\linewidth]{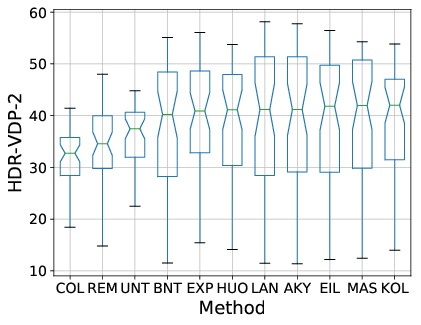}
    \end{minipage}
    \caption{Box plots for scene-referred HDR obtained from LDR via
    \textit{optimal} exposure.}\label{fig:boxplots_scene_optimal}
\end{figure}
\begin{figure}[t]
    \centering
    \begin{minipage}{0.48\linewidth}
        \centering
        \includegraphics[width=1.0\linewidth]{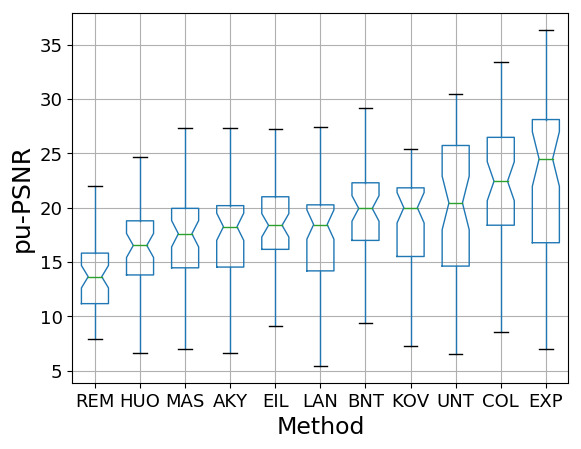}
    \end{minipage}
    \begin{minipage}{0.48\linewidth}
        \centering
        \includegraphics[width=1.0\linewidth]{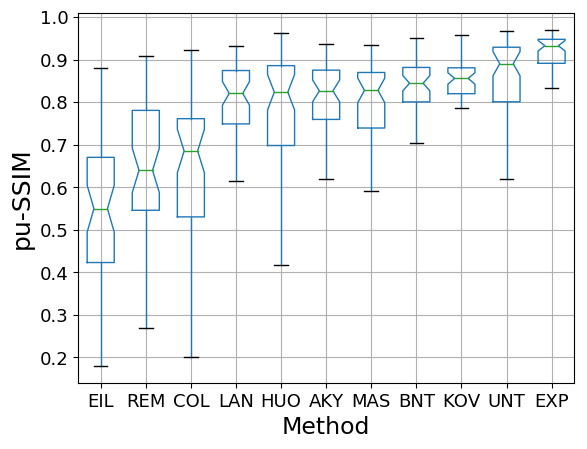}
    \end{minipage}\\
    \begin{minipage}{0.48\linewidth}
        \centering
        \includegraphics[width=1.0\linewidth]{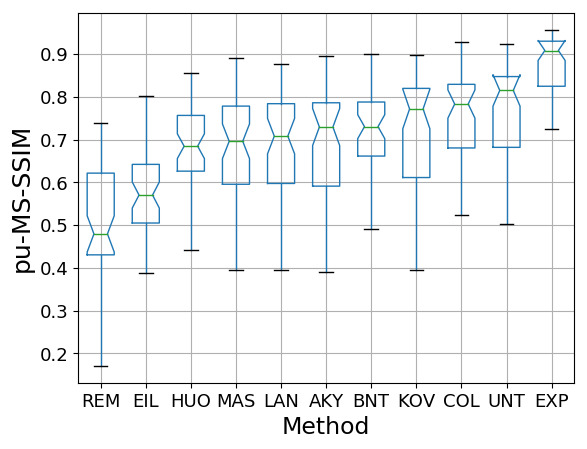}
    \end{minipage}
    \begin{minipage}{0.48\linewidth}
        \centering
        \includegraphics[width=1.0\linewidth]{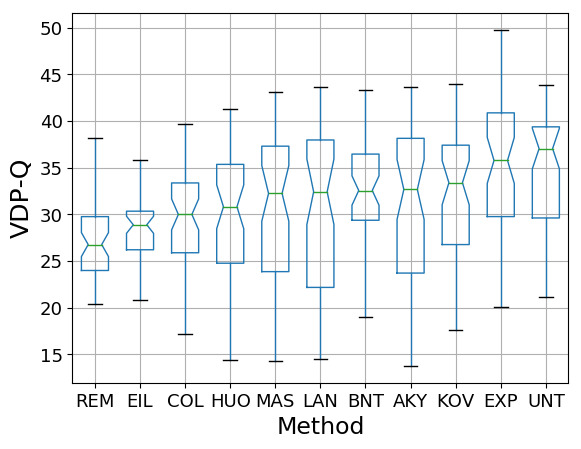}
    \end{minipage}
    \caption{Box plots for scene-referred HDR obtained from LDR via
    \textit{culling}.}\label{fig:boxplots_scene_culling}
\end{figure}
\begin{figure}[htb]
    \centering
    \begin{minipage}{0.48\linewidth}
    \centering
        \includegraphics[width=1.0\linewidth]{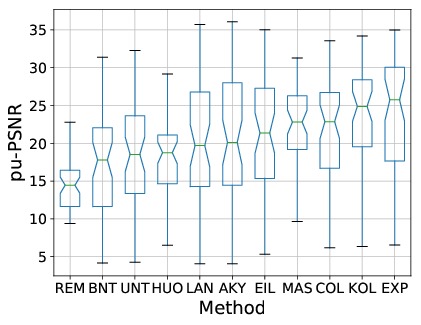}
    \end{minipage}
    \begin{minipage}{0.48\linewidth}
    \centering
        \includegraphics[width=1.0\linewidth]{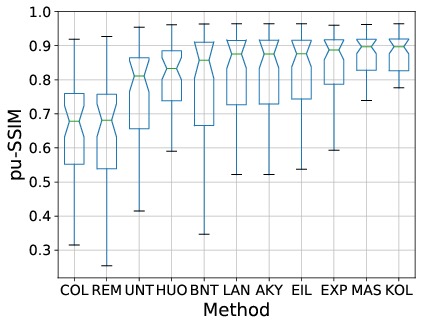}
    \end{minipage}\\
    \begin{minipage}{0.48\linewidth}
    \centering
        \includegraphics[width=1.0\linewidth]{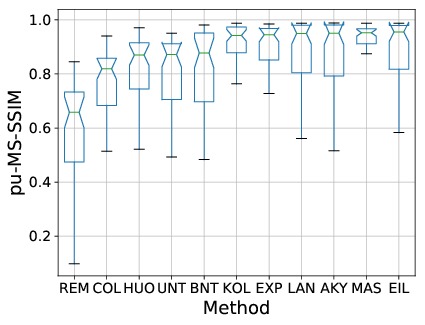}
    \end{minipage}
    \begin{minipage}{0.48\linewidth}
    \centering
        \includegraphics[width=1.0\linewidth]{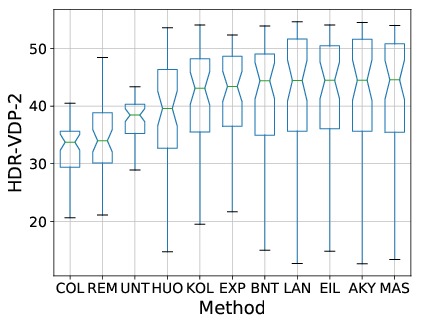}
    \end{minipage}
    \caption{Box plots for display-referred HDR obtained from LDR via
    \textit{optimal} exposure.}\label{fig:boxplots_tv_optimal}
\end{figure}
\begin{figure}[htb]
    \centering
    \begin{minipage}{0.48\linewidth}
    \centering
        \includegraphics[width=1.0\linewidth]{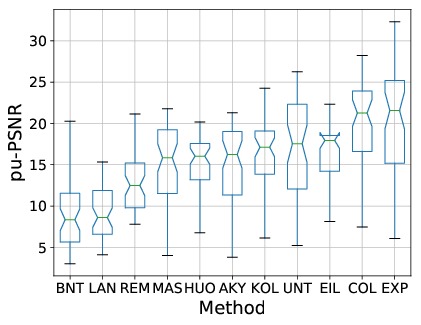}
    \end{minipage}
    \begin{minipage}{0.48\linewidth}
    \centering
        \includegraphics[width=1.0\linewidth]{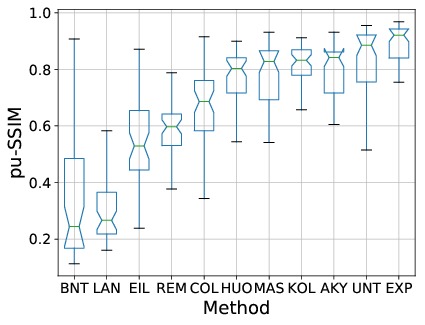}
    \end{minipage}\\
    \begin{minipage}{0.48\linewidth}
    \centering
        \includegraphics[width=1.0\linewidth]{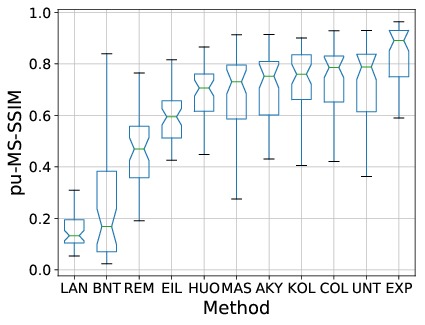}
    \end{minipage}
    \begin{minipage}{0.48\linewidth}
    \centering
        \includegraphics[width=1.0\linewidth]{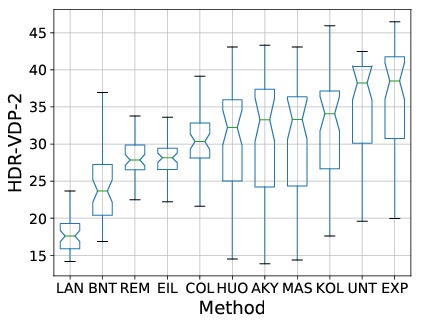}
    \end{minipage}
    \caption{Box plots for display-referred HDR obtained from LDR via
    \textit{culling}.}\label{fig:boxplots_tv_culling}
\end{figure}

\begin{figure*}[htb]
    \centering
    \begin{subfigure}[t]{0.13\linewidth}
        \centering
        \includegraphics[width=1.0\linewidth]{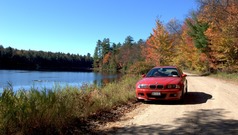}
        \caption{LDR}
    \end{subfigure}
    \begin{subfigure}[t]{0.13\linewidth}
        \centering
        \includegraphics[width=1.0\linewidth]{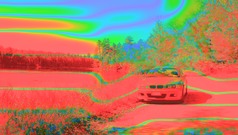}
        \caption{AKY}
    \end{subfigure}
    \begin{subfigure}[t]{0.13\linewidth}
        \centering
        \includegraphics[width=1.0\linewidth]{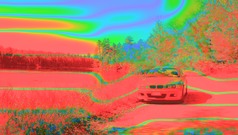}
        \caption{LAN}
    \end{subfigure}
    \begin{subfigure}[t]{0.13\linewidth}
        \centering
        \includegraphics[width=1.0\linewidth]{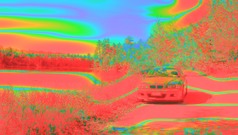}
        \caption{BNT}
    \end{subfigure}
    \begin{subfigure}[t]{0.13\linewidth}
        \centering
        \includegraphics[width=1.0\linewidth]{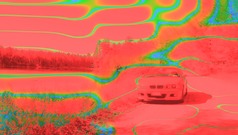}
        \caption{HUO}
    \end{subfigure}
    \begin{subfigure}[t]{0.13\linewidth}
        \centering
        \includegraphics[width=1.0\linewidth]{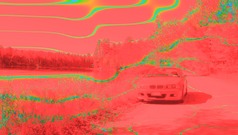}
        \caption{REM}
    \end{subfigure}\\
    \begin{subfigure}[t]{0.13\linewidth}
        \centering
        \includegraphics[width=1.0\linewidth]{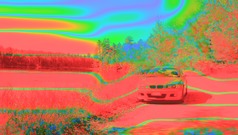}
        \caption{MAS}
    \end{subfigure}
    \begin{subfigure}[t]{0.13\linewidth}
        \centering
        \includegraphics[width=1.0\linewidth]{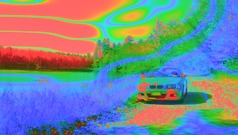}
        \caption{KOV}
    \end{subfigure}
    \begin{subfigure}[t]{0.13\linewidth}
        \centering
        \includegraphics[width=1.0\linewidth]{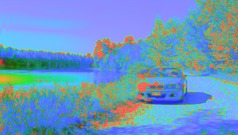}
        \caption{COL}
    \end{subfigure}
    \begin{subfigure}[t]{0.13\linewidth}
        \centering
        \includegraphics[width=1.0\linewidth]{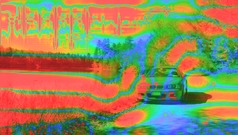}
        \caption{UNT}
    \end{subfigure}
    \begin{subfigure}[t]{0.13\linewidth}
        \centering
        \includegraphics[width=1.0\linewidth]{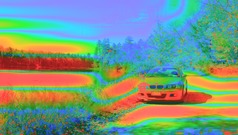}
        \caption{EIL}
    \end{subfigure}
    \begin{subfigure}[t]{0.13\linewidth}
        \centering
        \includegraphics[width=1.0\linewidth]{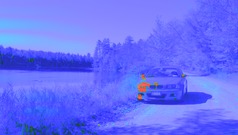}
        \caption{EXP}
    \end{subfigure}
    \caption{HDR-VDP-2.2 visibility probability maps for predictions of
    (\textit{culling}) M3 Middle Pond using all methods. Blue indicates
    imperceptible differences, red indicates perceptible
    differences.}\label{fig:mapsgood}
\end{figure*}
\begin{figure*}[htb]
    \centering
    \begin{subfigure}[t]{0.13\linewidth}
        \centering
        \includegraphics[width=1.0\linewidth]{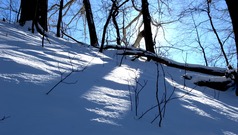}
        \caption{LDR}
    \end{subfigure}
    \begin{subfigure}[t]{0.13\linewidth}
        \centering
        \includegraphics[width=1.0\linewidth]{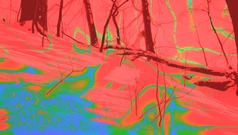}
        \caption{AKY}
    \end{subfigure}
    \begin{subfigure}[t]{0.13\linewidth}
        \centering
        \includegraphics[width=1.0\linewidth]{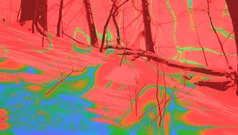}
        \caption{LAN}
    \end{subfigure}
    \begin{subfigure}[t]{0.13\linewidth}
        \centering
        \includegraphics[width=1.0\linewidth]{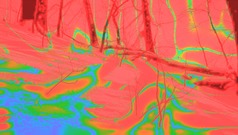}
        \caption{BNT}
    \end{subfigure}
    \begin{subfigure}[t]{0.13\linewidth}
        \centering
        \includegraphics[width=1.0\linewidth]{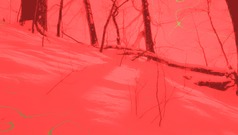}
        \caption{HUO}
    \end{subfigure}
    \begin{subfigure}[t]{0.13\linewidth}
        \centering
        \includegraphics[width=1.0\linewidth]{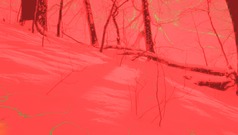}
        \caption{REM}
    \end{subfigure}\\
    \begin{subfigure}[t]{0.13\linewidth}
        \centering
        \includegraphics[width=1.0\linewidth]{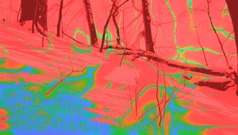}
        \caption{MAS}
    \end{subfigure}
    \begin{subfigure}[t]{0.13\linewidth}
        \centering
        \includegraphics[width=1.0\linewidth]{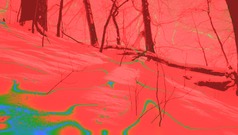}
        \caption{KOV}
    \end{subfigure}
    \begin{subfigure}[t]{0.13\linewidth}
        \centering
        \includegraphics[width=1.0\linewidth]{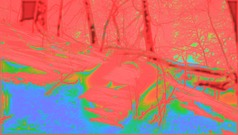}
        \caption{COL}
    \end{subfigure}
    \begin{subfigure}[t]{0.13\linewidth}
        \centering
        \includegraphics[width=1.0\linewidth]{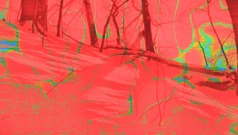}
        \caption{UNT}
    \end{subfigure}
    \begin{subfigure}[t]{0.13\linewidth}
        \centering
        \includegraphics[width=1.0\linewidth]{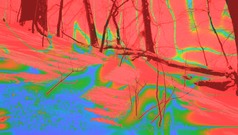}
        \caption{EIL}
    \end{subfigure}
    \begin{subfigure}[t]{0.13\linewidth}
        \centering
        \includegraphics[width=1.0\linewidth]{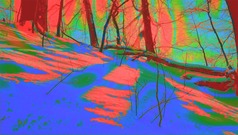}
        \caption{EXP}
    \end{subfigure}
    \caption{HDR-VDP-2.2 visibility probability maps for predictions of
    (\textit{culling}) Devils Bathtub using all methods. Blue indicates
    imperceptible differences, red indicates perceptible
    differences.}\label{fig:mapsaverage}
\end{figure*}

\begin{figure*}[htb]
    \centering
    \begin{subfigure}[t]{0.13\linewidth}
        \centering
        \includegraphics[width=1.0\linewidth]{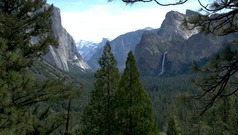}
        \caption{LDR}
    \end{subfigure}
    \begin{subfigure}[t]{0.13\linewidth}
        \centering
        \includegraphics[width=1.0\linewidth]{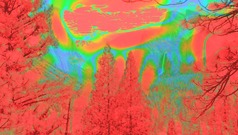}
        \caption{AKY}
    \end{subfigure}
    \begin{subfigure}[t]{0.13\linewidth}
        \centering
        \includegraphics[width=1.0\linewidth]{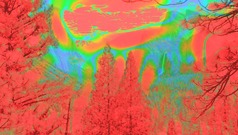}
        \caption{LAN}
    \end{subfigure}
    \begin{subfigure}[t]{0.13\linewidth}
        \centering
        \includegraphics[width=1.0\linewidth]{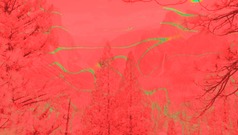}
        \caption{BNT}
    \end{subfigure}
    \begin{subfigure}[t]{0.13\linewidth}
        \centering
        \includegraphics[width=1.0\linewidth]{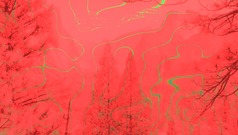}
        \caption{HUO}
    \end{subfigure}
    \begin{subfigure}[t]{0.13\linewidth}
        \centering
        \includegraphics[width=1.0\linewidth]{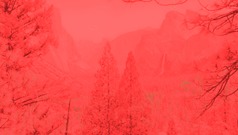}
        \caption{REM}
    \end{subfigure}\\
    \begin{subfigure}[t]{0.13\linewidth}
        \centering
        \includegraphics[width=1.0\linewidth]{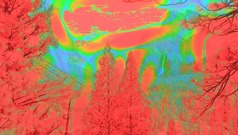}
        \caption{MAS}
    \end{subfigure}
    \begin{subfigure}[t]{0.13\linewidth}
        \centering
        \includegraphics[width=1.0\linewidth]{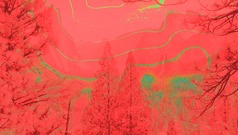}
        \caption{KOV}
    \end{subfigure}
    \begin{subfigure}[t]{0.13\linewidth}
        \centering
        \includegraphics[width=1.0\linewidth]{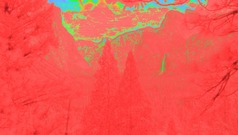}
        \caption{COL}
    \end{subfigure}
    \begin{subfigure}[t]{0.13\linewidth}
        \centering
        \includegraphics[width=1.0\linewidth]{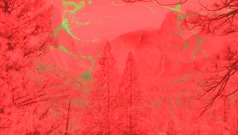}
        \caption{UNT}
    \end{subfigure}
    \begin{subfigure}[t]{0.13\linewidth}
        \centering
        \includegraphics[width=1.0\linewidth]{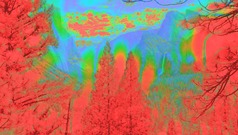}
        \caption{EIL}
    \end{subfigure}
    \begin{subfigure}[t]{0.13\linewidth}
        \centering
        \includegraphics[width=1.0\linewidth]{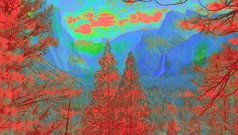}
        \caption{EXP}
    \end{subfigure}
    \caption{HDR-VDP-2.2 visibility probability maps for predictions of
    (\textit{culling}) Tunnel View using all methods. Blue indicates
    imperceptible differences, red indicates perceptible
    differences.}\label{fig:mapsbad}
\end{figure*}

\begin{figure*}[htb]
    \centering
    \begin{subfigure}[t]{0.27\linewidth}
        \centering
        \includegraphics[width=1.0\linewidth]{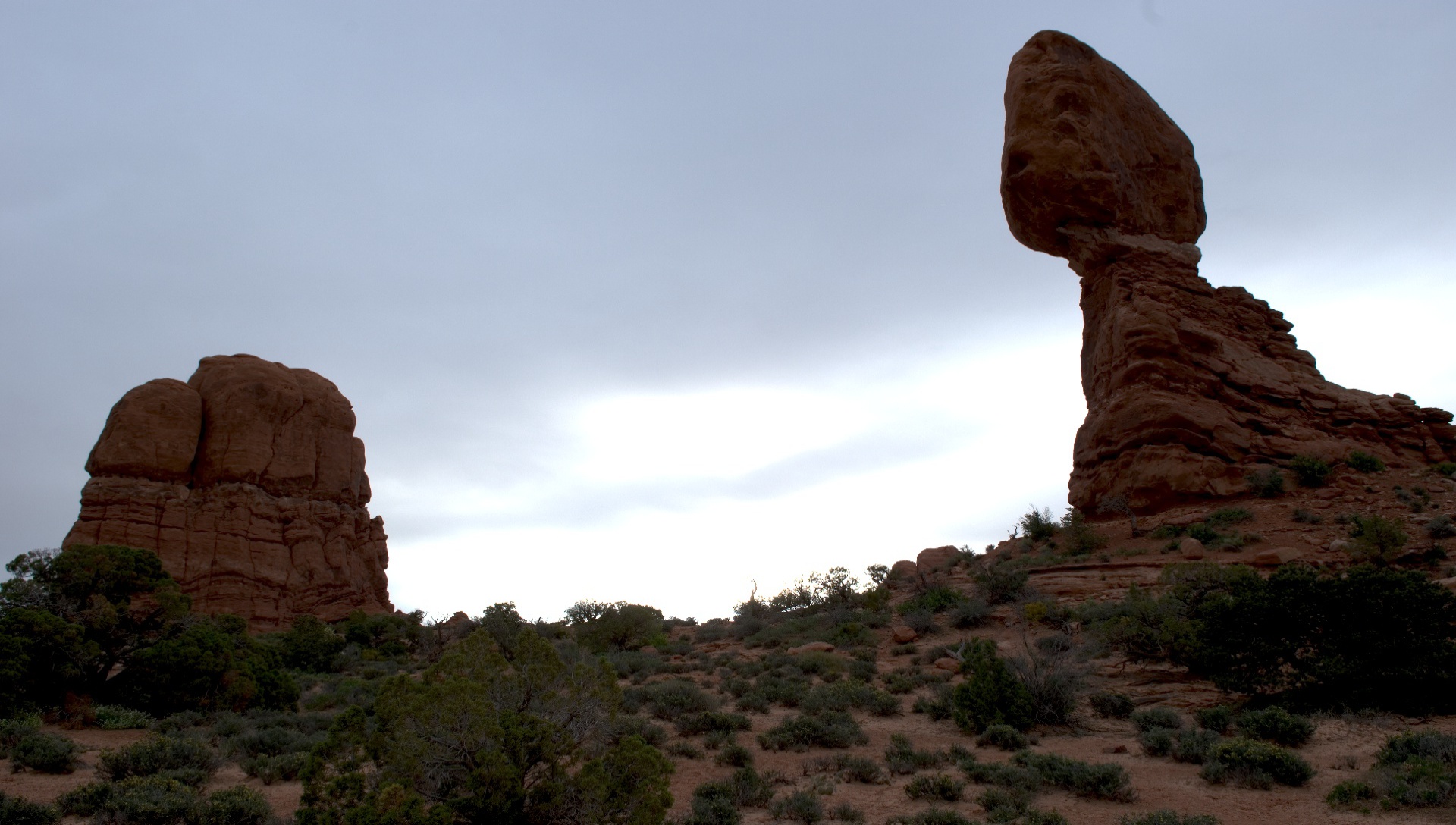}
        \caption{Input LDR (culling)}
    \end{subfigure}
    \begin{subfigure}[t]{0.27\linewidth}
        \centering
        \includegraphics[width=1.0\linewidth]{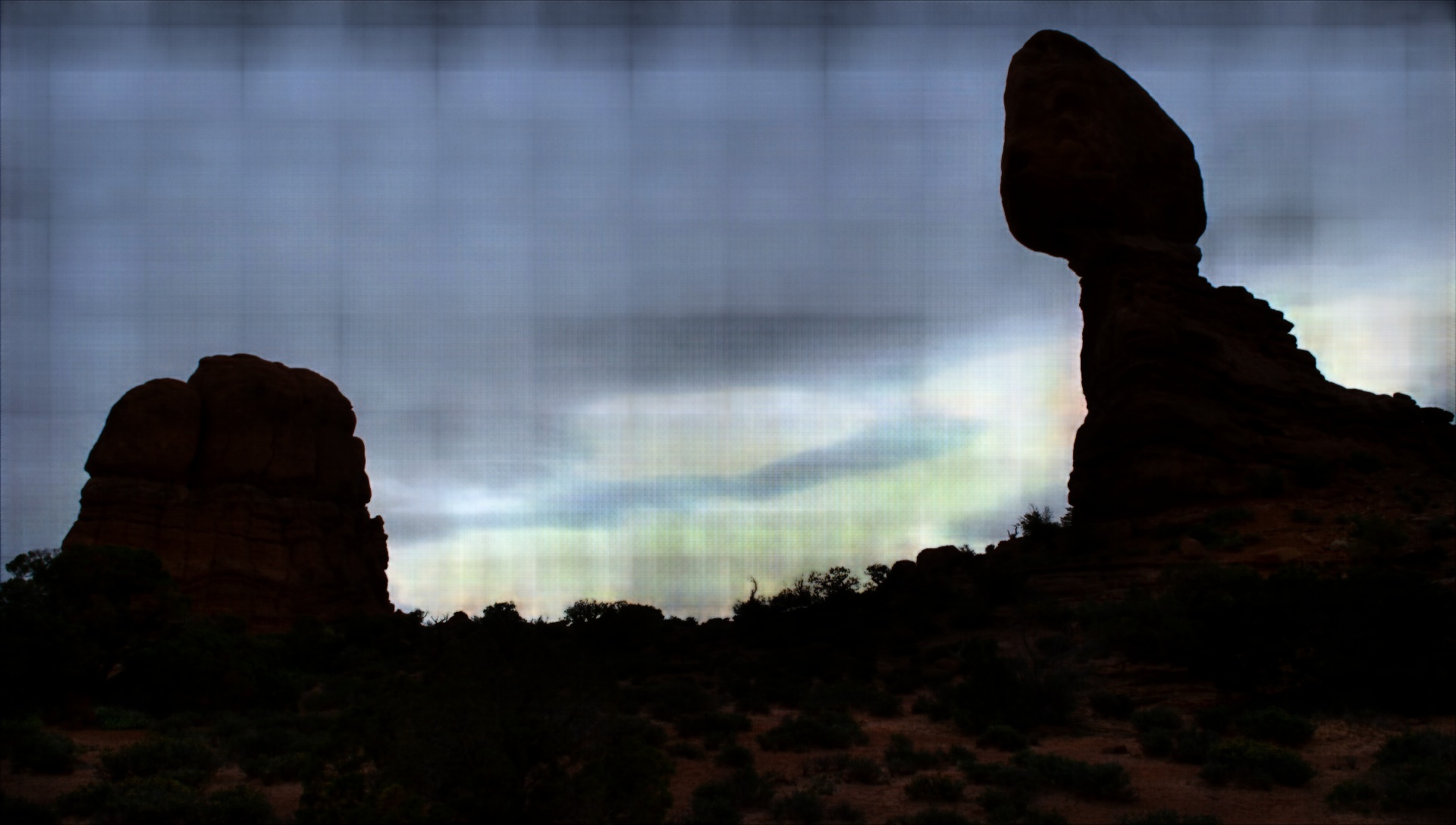}
        \caption{UNT}
    \end{subfigure}
    \begin{subfigure}[t]{0.27\linewidth}
        \centering
        \includegraphics[width=1.0\linewidth]{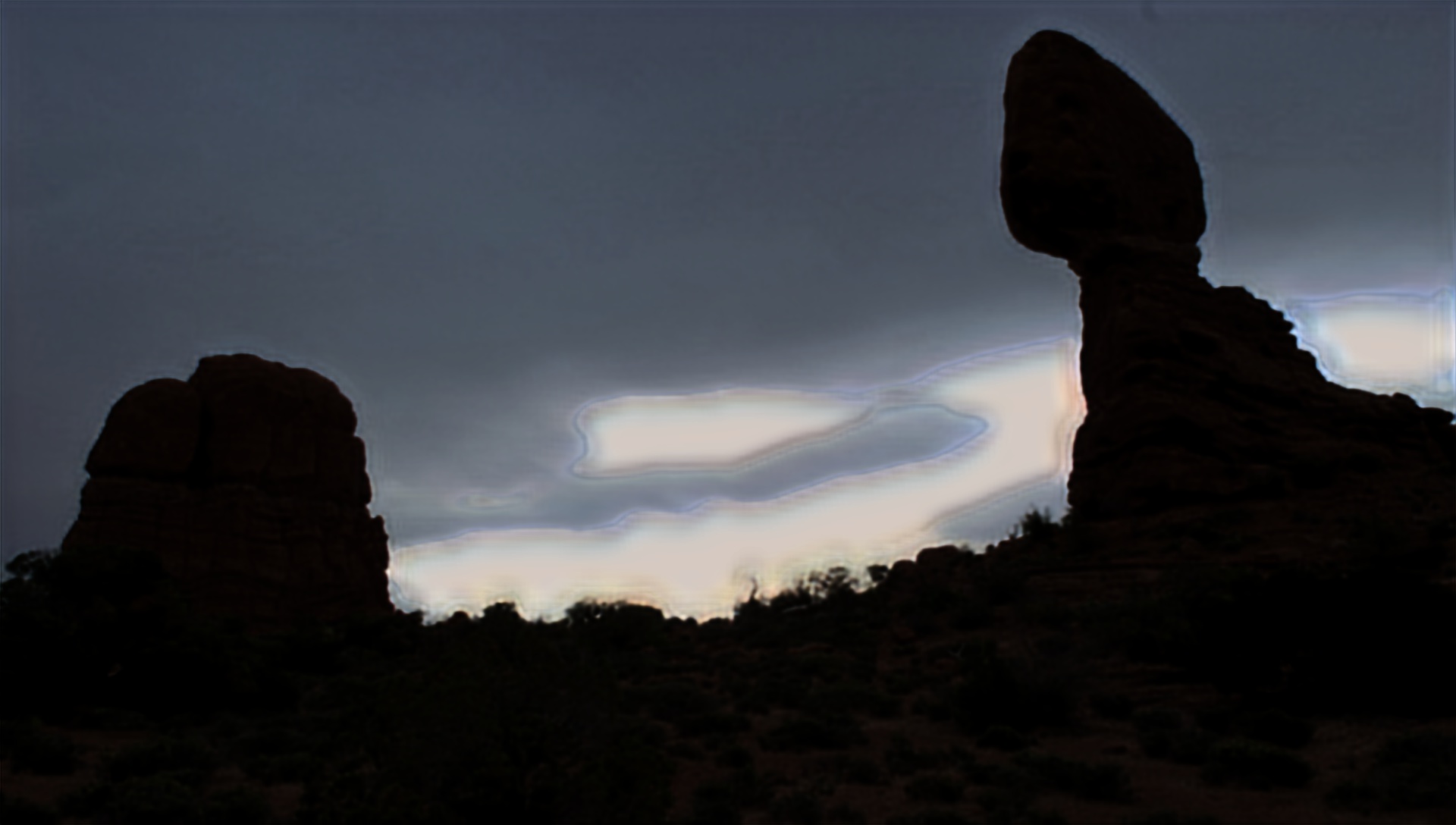}
        \caption{COL}
    \end{subfigure}
    \begin{subfigure}[t]{0.27\linewidth}
        \centering
        \includegraphics[width=1.0\linewidth]{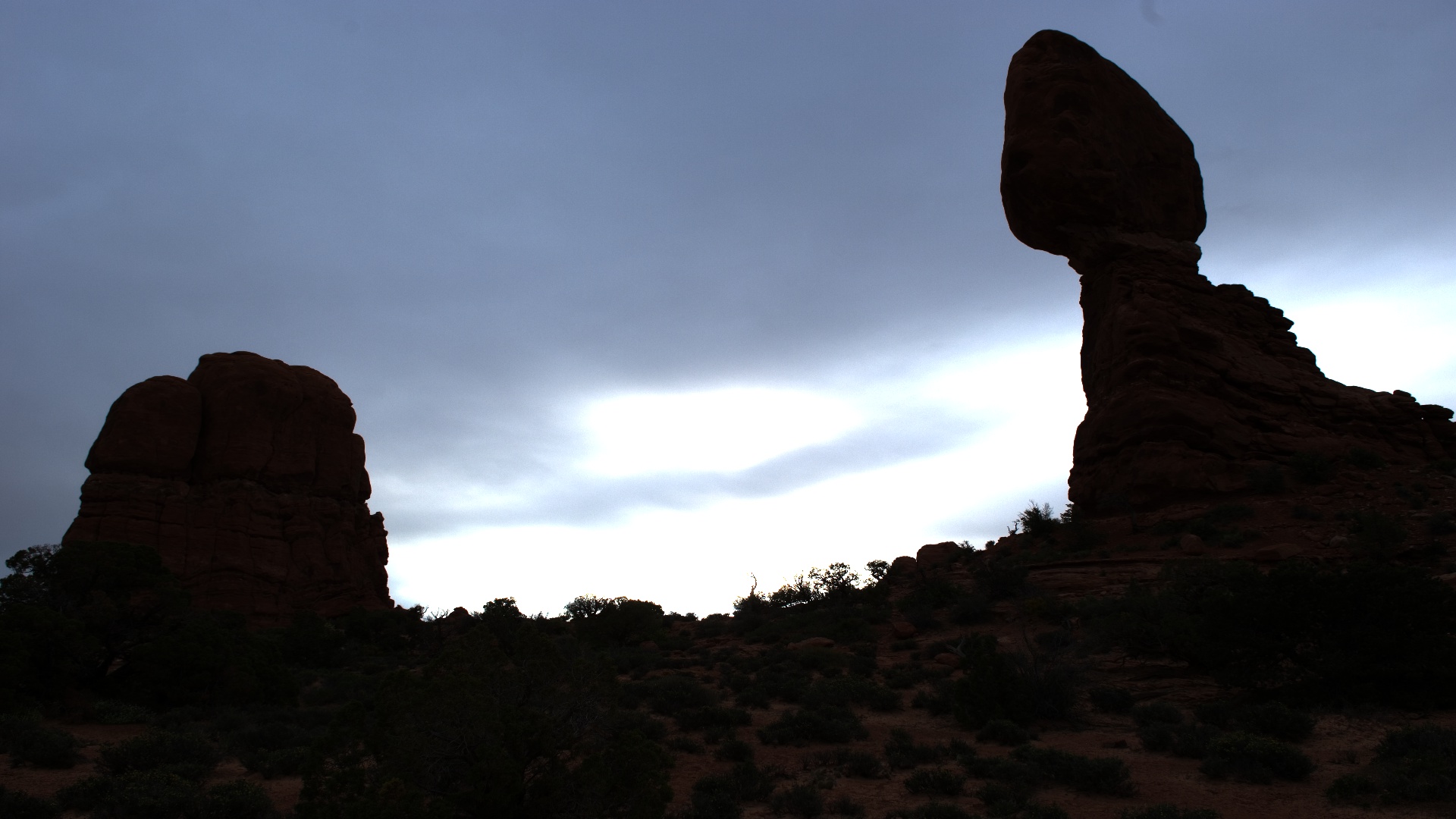}
        \caption{Exposure of original HDR}
    \end{subfigure}
    \begin{subfigure}[t]{0.27\linewidth}
        \centering
        \includegraphics[width=1.0\linewidth]{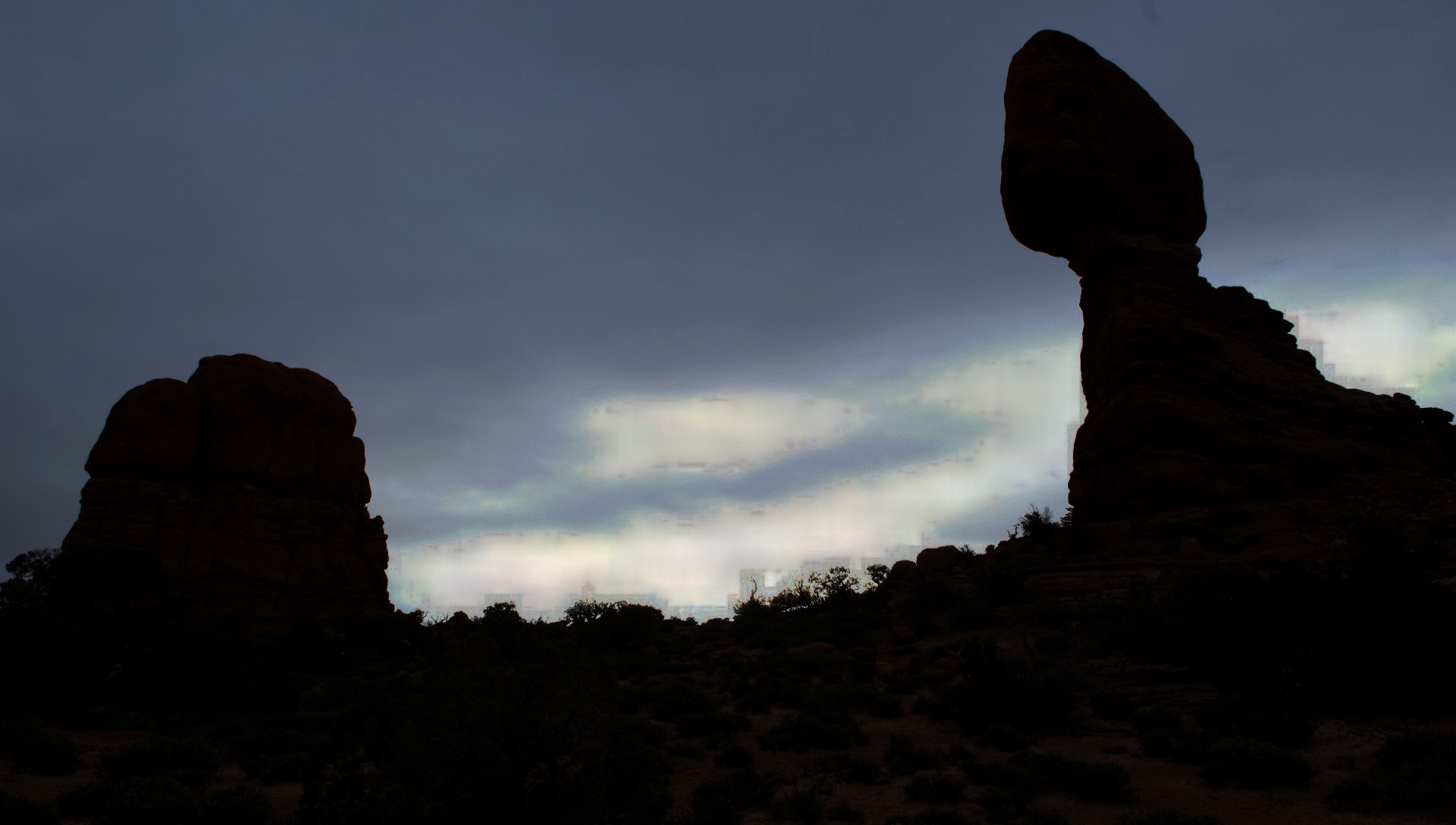}
        \caption{EIL}
    \end{subfigure}
    \begin{subfigure}[t]{0.27\linewidth}
        \centering
        \includegraphics[width=1.0\linewidth]{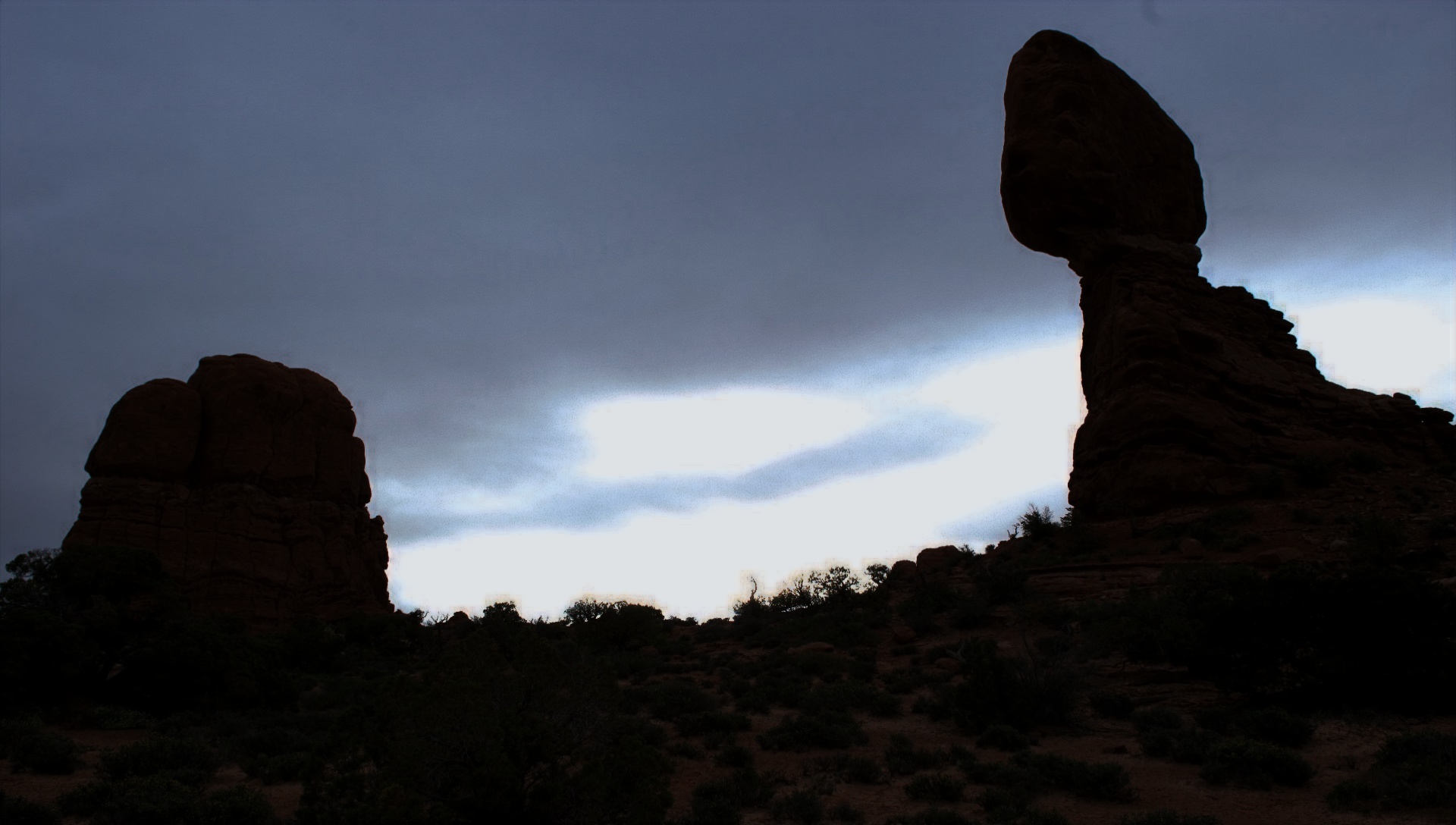}
        \caption{EXP}
    \end{subfigure}
    \caption{(a) LDR input image created using culling from the Balanced Rock HDR image. (d) Low exposure of the
    original HDR image.\\(b,c,e,f) Low exposure slices of the predictions from methods that use CNN architectures
    showing artefacts.}\label{fig:hallucinations_low}
\end{figure*} 
\begin{figure*}[htb]
    \centering
    \begin{subfigure}[t]{0.28\linewidth}
        \centering
        \includegraphics[width=1.0\linewidth]{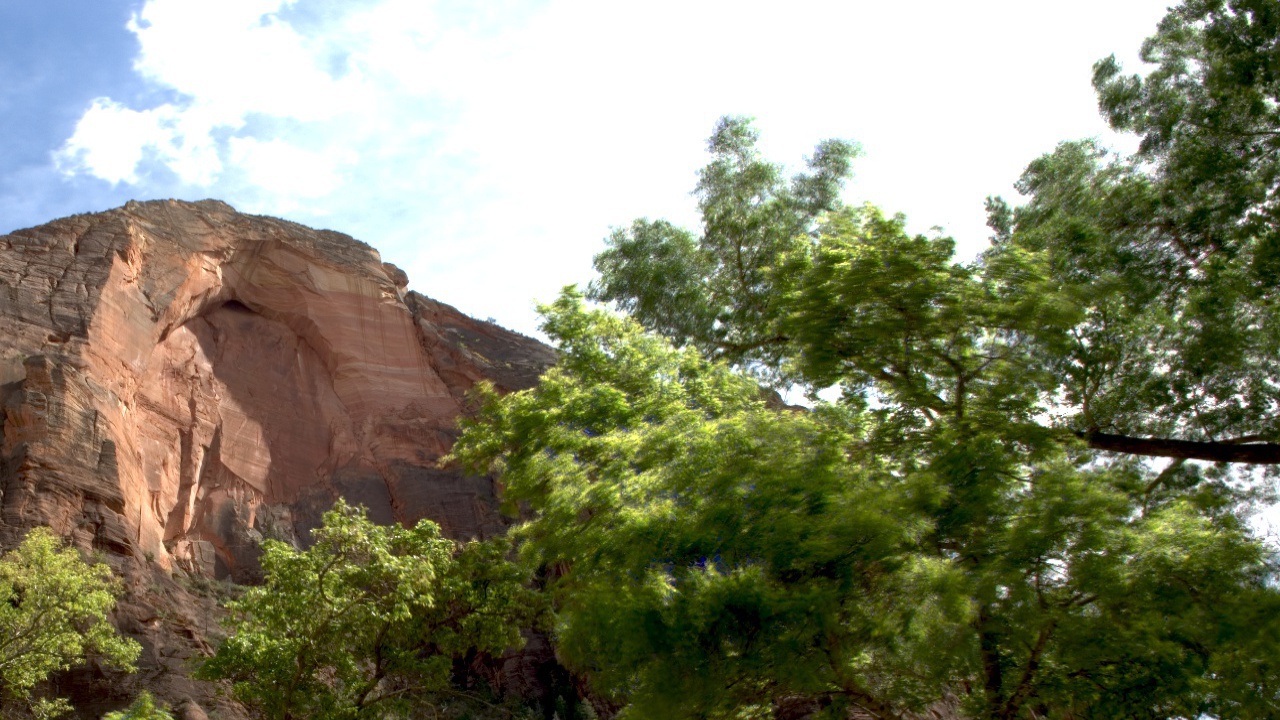}
        \caption{Input LDR (culling)}
    \end{subfigure}
    \begin{subfigure}[t]{0.28\linewidth}
        \centering
        \includegraphics[width=1.0\linewidth]{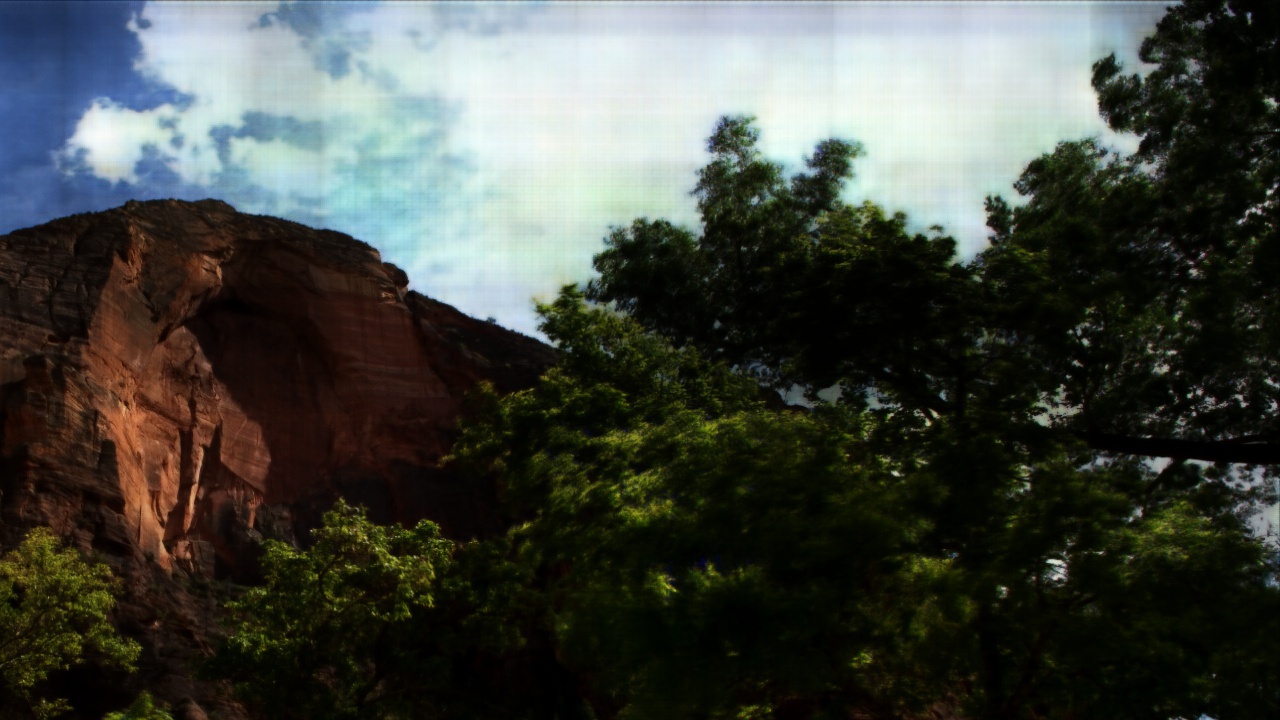}
        \caption{UNT}
    \end{subfigure}
    \begin{subfigure}[t]{0.28\linewidth}
        \centering
        \includegraphics[width=1.0\linewidth]{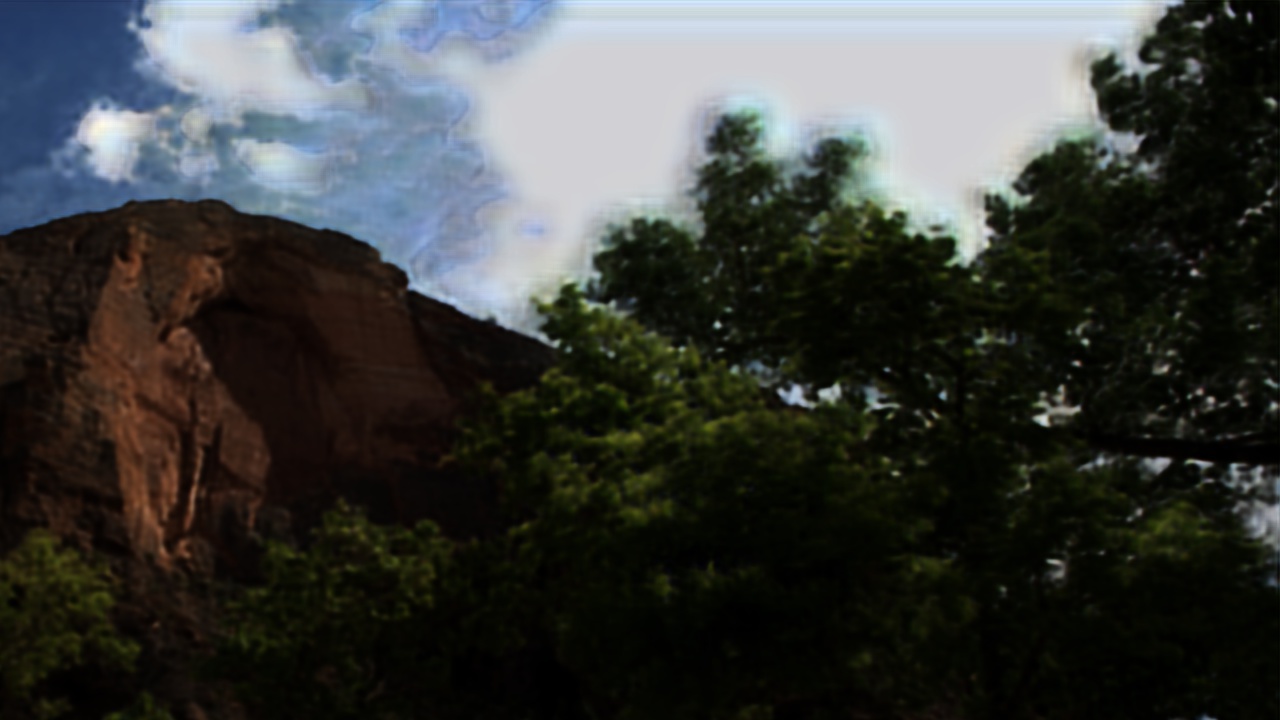}
        \caption{COL}
    \end{subfigure}
    \begin{subfigure}[t]{0.28\linewidth}
        \centering
        \includegraphics[width=1.0\linewidth]{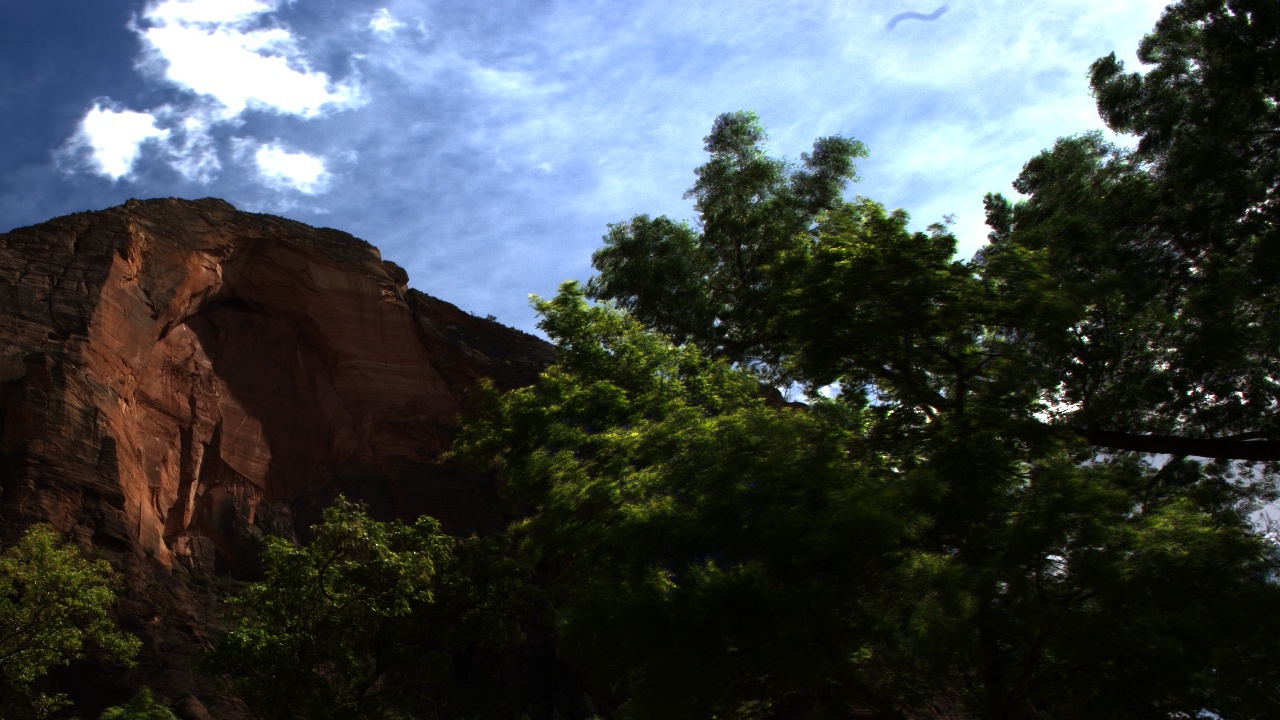}
        \caption{Exposure of original HDR}
    \end{subfigure}
    \begin{subfigure}[t]{0.28\linewidth}
        \centering
        \includegraphics[width=1.0\linewidth]{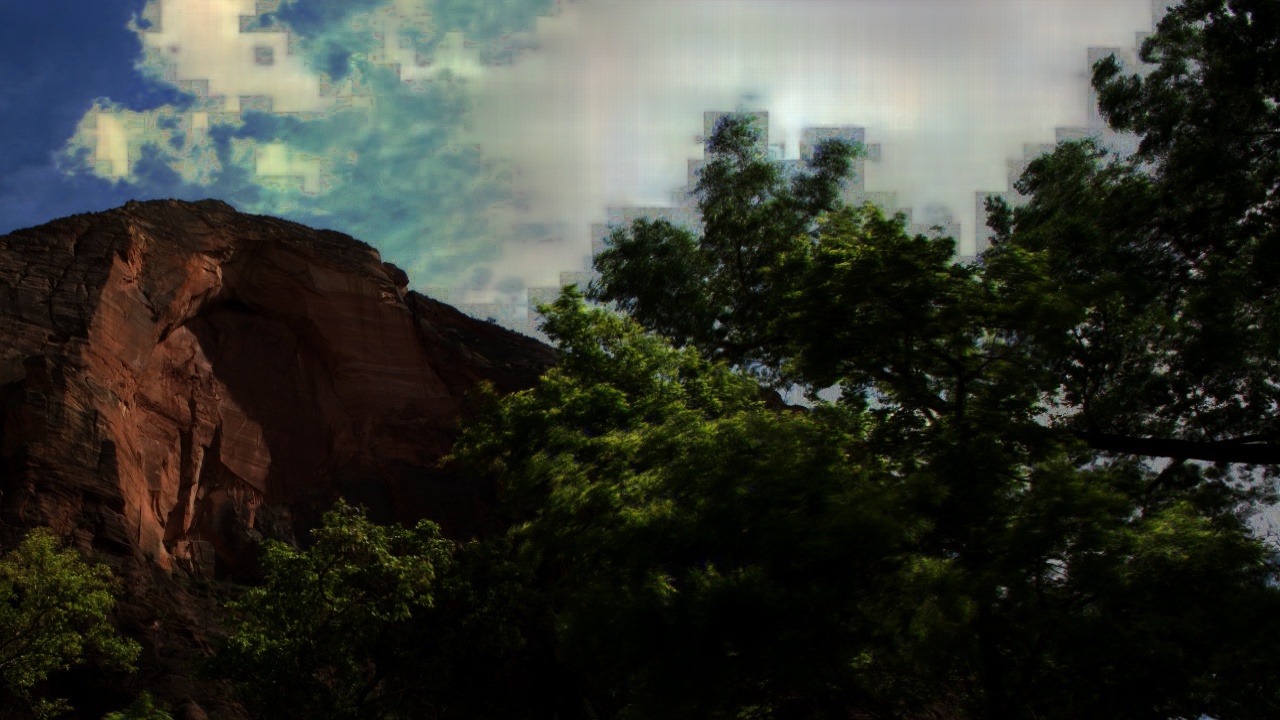}
        \caption{EIL}
    \end{subfigure}
    \begin{subfigure}[t]{0.28\linewidth}
        \centering
        \includegraphics[width=1.0\linewidth]{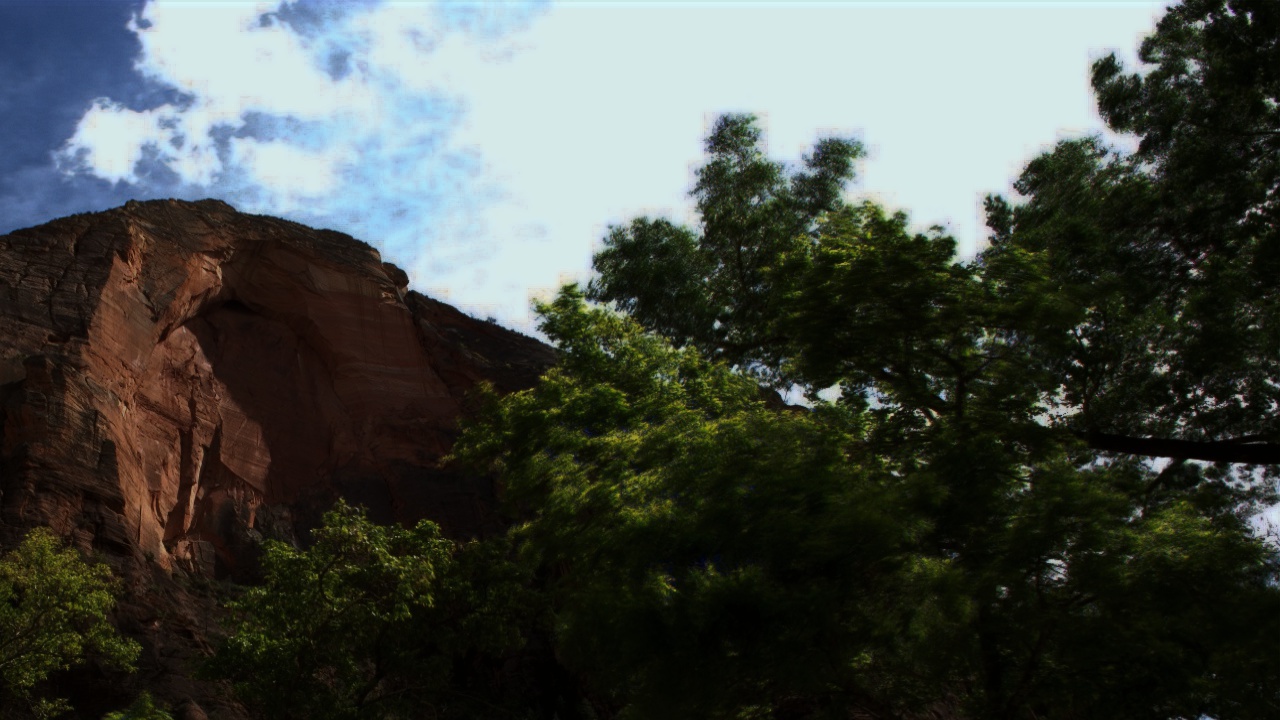}
        \caption{EXP}
    \end{subfigure}
    \caption{(a) LDR input image created using culling from The Grotto HDR
    image. (d) Low exposure of the original HDR image.\\(b,c,e,f) Low exposure
    slices of the predictions from methods that use CNN architectures showing
    artefacts.}\label{fig:hallucinations_grotto}
\end{figure*} 

\subsection{\textbf{Dataset}}\label{sec:dataset}

A dataset of HDR images was created consisting of 1,013 training images and 50
test images, with resolutions ranging from $800\times800$ up to
$4,916\times3,273$. The images were collected from various sources, including
in-house images, frames from HDR videos and the web. Only 100 of the images
contained calibrated luminance values, sourced from the Fairchild database
\cite{fairchild2007hdr}. All the images contained linear RGB values. The 50
test images used for evaluation in Section~\ref{sec:results} were selected
randomly from the Fairchild images with calibrated absolute luminance. LDR
content for training was generated on-the-fly, directly from the dataset, and
was augmented in a number of ways as outlined below.

At every epoch each HDR image from the training set is used as input in the
network once after preprocessing. Preprocessing consists of randomly selecting
a position for a sub image, cropping, and having its dynamic range reduced
using one of a set of operators. The randomness entails that at every epoch a
different LDR-HDR pair is generated from a single HDR image in the training
set.

Initially, the HDR image has its cropping position selected. The position is drawn from a spatial Gaussian distribution such that the most frequently
selected regions are towards the center of the image. The crop size is drawn from an exponential distribution such that smaller crops are more
frequent than larger ones, with a minimum crop size of $384\times384$. Randomly cropping the images is a standard technique for data augmentation.
Choosing the crop size at random adds another layer of augmentation, since the likelihood of picking the same crop is reduced, but it also aids in
how well the model generalizes since it provides different sized content for similar scenes.

The cropped image is resized to $256\times256$ and linearly mapped to the
$[0,1]$ range to create the output. Since only a small fraction of the dataset
images contain absolute luminance values, the network was trained to predict
relative luminance values in the $[0,1]$ range.

A tone mapping operator (TMO)~\cite{tumblin1993tonemap} or single exposure operator is applied to form the input LDR from the output HDR, chosen
uniformly from a list of five operators: dynamic range reduction inspired by photoreceptor physiology (Photoreceptor)~\cite{reinhard2005dynamic},
Adaptive Logarithmic Mapping (ALM)~\cite{drago2003tmo}, Display Adaptive Tone Mapping (display)~\cite{mantiuk08display},
Bilateral~\cite{durand2002tmo} and Exposure. The OpenCV3 implementations of the TMOs were used. The Exposure operator was implemented for this work
and clamps the top and bottom percentiles of the image and adds a gamma curve. In addition to using a random operator for each input-output pair, the
parameters of the operators are also randomized. The parameters of the functions used are summarized in Table~\ref{table:tmo_param}. The TMO
parameter randomization was done to ensure that the model performs well under a variety of inputs when tested with real LDR inputs and does not just
learn to invert specific TMOs. It acts as yet another layer of data augmentation. Results shown in the following section only use single exposures
for generating HDR; the TMOs are just used for data augmentation during training.

\subsection{\textbf{Optimization}}

The network parameters are optimized to minimize the loss given in
Equation~\ref{eq:Loss}, with $\lambda=5$, using mini-batch gradient descent and
the backpropagation algorithm~\cite{rumelhart1986learning}. The Adam optimizer
was used~\cite{kingma2014adam}, with an initial learning rate of $7e-5$ and a
batch size of $12$. After the first $10,000$ epochs, the learning rate was
reduced by a factor of $0.8$ whenever the loss reached a plateau, until the
learning rate reached values less than $1e-7$ for a total of $1,600$ epochs
extra. $L_2$ regularization (weight decay) was used to reduce the chance of
overfitting. All experiments were implemented using the PyTorch
library~\cite{pytorch}. Training time took a total of 130 hours on an Nvidia
P100.

\section{\textbf{Results}}
\label{sec:results}

This section presents an evaluation of ExpandNet compared to other EOs and deep
learning architectures. Figure~\ref{fig:workflow} (right) shows an overview of
the evaluation method.

\subsection{\textbf{Quantitative}}

For a quantitative evaluation of the work, four metrics are considered, Peak
Signal to Noise Ratio (PSNR), Structural Similarity (SSIM), Multi-Scale
Structural Similarity (MS-SSIM), and  HDR-VDP-2.2~\cite{narwaria2015hdrvdp}.
For the first three metrics, a perceptual uniformity (PU)
encoding~\cite{aydin2008} is applied to the prediction and reference images to
make them suitable for HDR comparisons. HDR-VDP-2.2 includes the PU-encoding in
its implementation. The values from HDR-VDP-2.2 are those of the VDP-Q quality
score.

The ExpandNet architecture is compared against seven other previous methods for
dynamic range expansion/inverse tone mapping. The chosen methods were the
methods of: Landis~\cite{Landis02} (LAN), Banterle et
al.~\cite{banterle06inverse} (BNT), Aky\"{u}z et al.~\cite{akyuz06} (AKY),
Rempel et al.~\cite{rempel06ldr2hdr} (REM),  Masia et al.~\cite{masia09}
(MAS), Kovaleski and Oliveira~\cite{Kovaleski+2014} (KOV) and Huo et
al.~\cite{Huo+2014} (HUO). The Matlab implementations from the HDR
toolbox~\cite{banterle2011hdrbook} were used to obtain these results.

Four CNN architectures are compared, including the proposed ExpandNet method
(EXP). Two other network architectures that have been used for
similar problems have been adopted and trained in the same way as EXP. The first
network is based on U-Net~\cite{ronneberger2015unet} (UNT), an architecture
that has shown strong results with image translation tasks between domains. The
second network is an architecture first used for
colorization~\cite{iizuka2016colornet} (COL), which uses two branches and a
fusion layer similar to the one used for ExpandNet. These three are implemented
using the same pyTorch framework and trained on the same training dataset. The
recent network architecture used for LDR to HDR
conversion~\cite{eilertsen2017cnn} (EIL) is also included. The predictions
from this method were created using the trained network which was made
available online by the authors, applied on the same test dataset used for all
the other methods.

The inputs to the methods are single exposure LDR images of the 50 full HD
($1920 \times 1080$) images in the HDR test dataset. The single exposures are
obtained using two methods. The first method (\textit{optimal}) finds the
optimal/automatic exposure~\cite{debattista2015optimal} using the HDR image
histogram, resulting in minimal clipping at the two ends of the luminance
range. The second method (\textit{culling}) simply clips the top and bottom
$10\%$ of the values of the images, resulting in more information loss and
distortion of the input LDR. The resulting test LDR input images are saved with
JPEG encoding before testing. When compared to the $10^{\text{th}}$ percentile
loss for the images generated using \textit{culling}, on average, the number of
pixels over the test dataset that are over-exposed when using \textit{optimal}
is 3.89\% and the number of pixels under-exposed is 0.35\%.

The outputs of the methods are in the $[0, 1]$ range, predicting relative
luminance. The scaling permits evaluation for scene-referred and
display-referred output. Hence, the predicted HDR images are scaled to match
the original HDR content (scene-referred) and a 1,000 $cd/m^2$ display
(display-referred), which represents current commercial HDR display technology.
The scaling is done to match the $0.1$ and $99.9$ percentiles of the
predictions with the corresponding percentiles of the HDR test images.
Furthermore, scaling is useful as the PU-encoded HDR metrics are dependent on
absolute luminance values in $cd/m^2$. By scaling the prediction outputs, the
PU-encoded metrics can be used to quantify the ability of the network to
reconstruct the original signal.

Table~\ref{table:resultsscene} and Table~\ref{table:resultsdisplay} summarize
the results of the four metrics applied on all the methods, using the
\textit{optimal} and \textit{culling}, for scene-referred and display-referred
respectively. Box plots for the distribution of the four metrics are presented
in Figure~\ref{fig:boxplots_scene_optimal} and
Figure~\ref{fig:boxplots_scene_culling} for the scene-referred results of
\textit{optimal} and \textit{culling} respectively. Similarly
Figure~\ref{fig:boxplots_tv_optimal} and Figure~\ref{fig:boxplots_tv_culling}
show the display-referred results of \textit{optimal} and \textit{culling}
respectively. Box plots are sorted by ascending order of median value.
\tc{When analysed for significant differences amongst all the methods, a
significance is found for all tests (at p < 0.001) using Friedman's test.
Pairwise comparisons ranked EXP in the top group, consisting of the group of
methods that cannot be significantly differentiated, in 13 of the 16 results
(these consist of all four metrics for both \textit{optimal} and
\textit{culling} and for both scene-referred and display referred). The
conditions where EXP was not in the top group were: pu-SSIM (in the cases of
scene-referred and display-referred) and pu-MMSIM (for scene-referred only); in
all three cases this occurred for the \textit{optimal} condition.}

As can be seen in the overall, EXP performs reasonably well. In particular for
the \textit{culling} case when a significant number of pixels are over or
under-exposed EXP appears to reproduce HDR better than the other methods. For
\textit{optimal}, EIL performs very well also, and this is expected as in such
cases the number of pixels that are required to be predicted from the CNN are
smaller. Similarly, the non deep learning based expansion methods such as MAS
perform well especially for SSIM which quantifies structural similarity.

\subsection{\textbf{Visual Inspection}}\label{sec:comparemodels}

This section presents some qualitative aspects of the results. HDR-VDP-2.2
visibility probability maps for all the methods are presented, as well as
images from the CNN predictions exhibiting effects such as hallucination,
blocking and information bleeding.

Figure~\ref{fig:mapsgood}, Figure~\ref{fig:mapsaverage} and
Figure~\ref{fig:mapsbad} show the HDR-VDP-2.2 probability maps for the
predictions of all the methods from the test set. The HDRs are predicted from
\textit{culling} LDRs with scene-referred scaling. The HDR-VDP-2.2 visibility
probability map describes how likely it is for a difference to be noticed by
the average observer, at each pixel. Red values indicate high probability,
while blue values indicate low probability. Results show EXP performs better
than most other methods for these scenes. EIL also performs well, particularly
for the challenging scenario in Figure~\ref{fig:mapsbad}.

\begin{figure*}[htb]
    \centering
    \begin{subfigure}[t]{0.40\linewidth}
        \centering
        \includegraphics[width=0.325\linewidth]{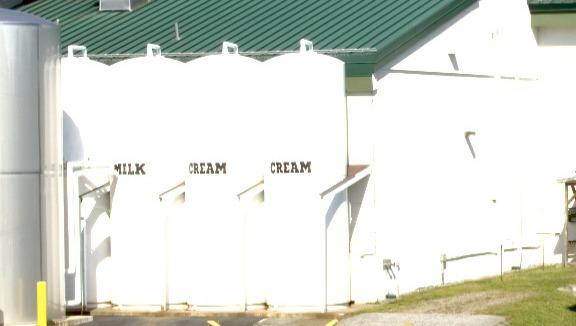}
    \end{subfigure}
    \begin{subfigure}[t]{0.40\linewidth}
        \centering
        \includegraphics[width=0.325\linewidth]{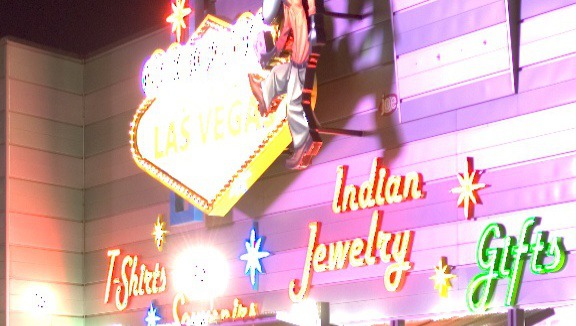}
    \end{subfigure}

    \begin{subfigure}[t]{0.13\linewidth}
        \centering
        \includegraphics[width=1.0\linewidth]{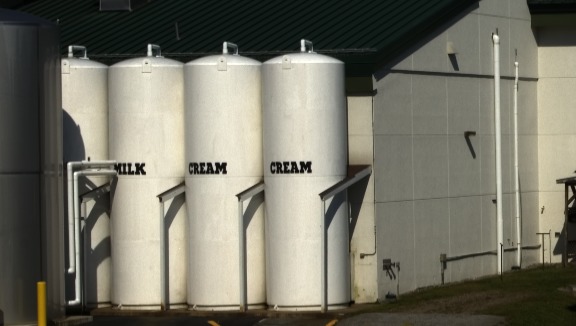}
    \end{subfigure}
    \begin{subfigure}[t]{0.13\linewidth}
        \centering
        \includegraphics[width=1.0\linewidth]{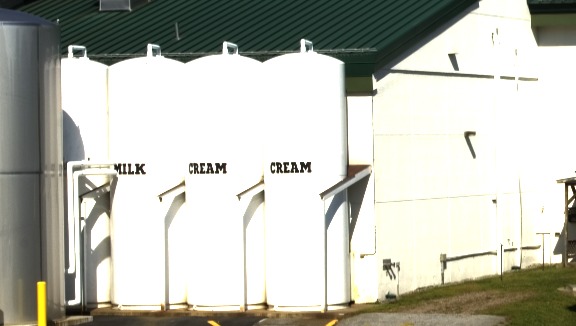}
    \end{subfigure}
    \begin{subfigure}[t]{0.13\linewidth}
        \centering
        \includegraphics[width=1.0\linewidth]{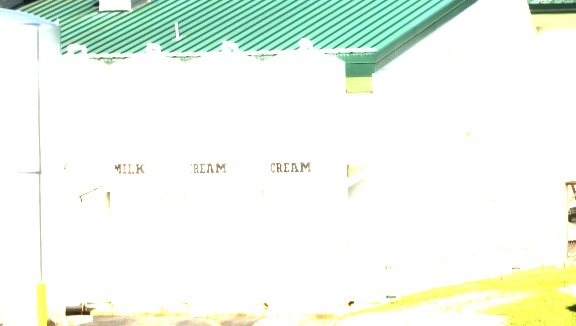}
    \end{subfigure}
    \begin{subfigure}[t]{0.13\linewidth}
        \centering
        \includegraphics[width=1.0\linewidth]{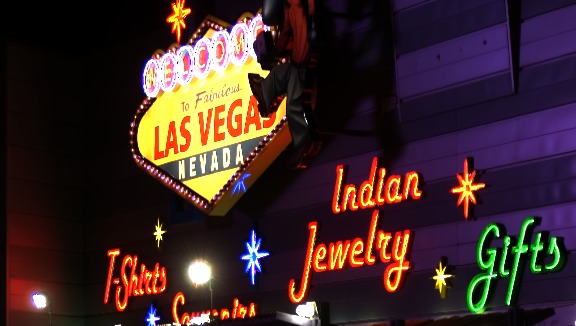}
    \end{subfigure}
    \begin{subfigure}[t]{0.13\linewidth}
        \centering
        \includegraphics[width=1.0\linewidth]{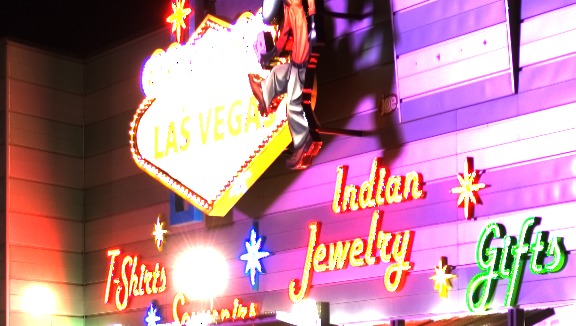}
    \end{subfigure}
    \begin{subfigure}[t]{0.13\linewidth}
        \centering
        \includegraphics[width=1.0\linewidth]{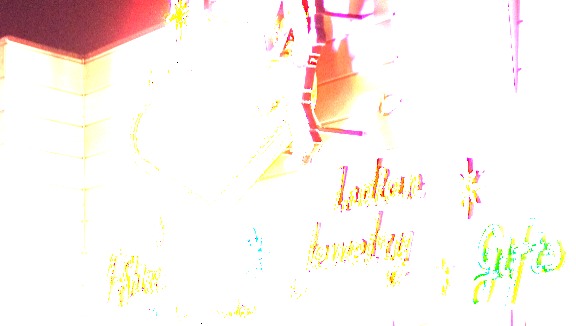}
    \end{subfigure}

    \begin{subfigure}[t]{0.13\linewidth}
        \centering
        \includegraphics[width=1.0\linewidth]{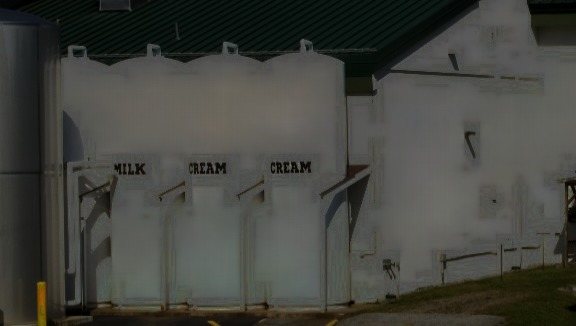}
    \end{subfigure}
    \begin{subfigure}[t]{0.13\linewidth}
        \centering
        \includegraphics[width=1.0\linewidth]{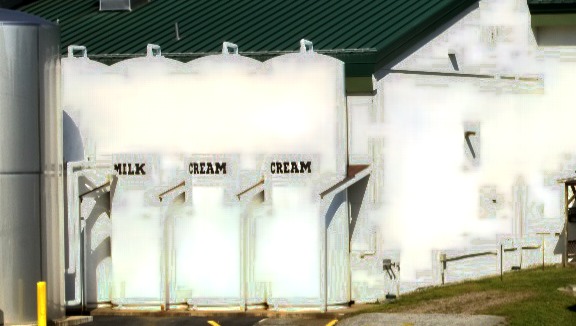}
    \end{subfigure}
    \begin{subfigure}[t]{0.13\linewidth}
        \centering
        \includegraphics[width=1.0\linewidth]{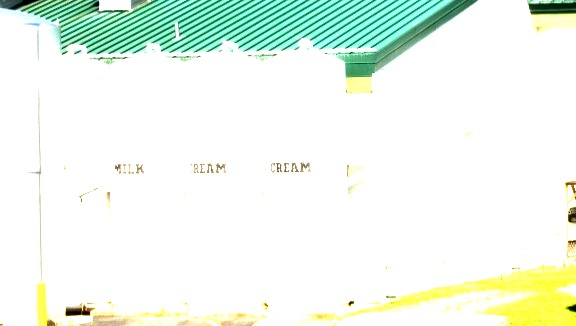}
    \end{subfigure}
    \begin{subfigure}[t]{0.13\linewidth}
        \centering
        \includegraphics[width=1.0\linewidth]{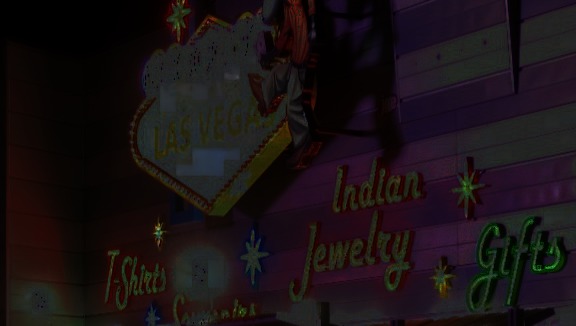}
    \end{subfigure}
    \begin{subfigure}[t]{0.13\linewidth}
        \centering
        \includegraphics[width=1.0\linewidth]{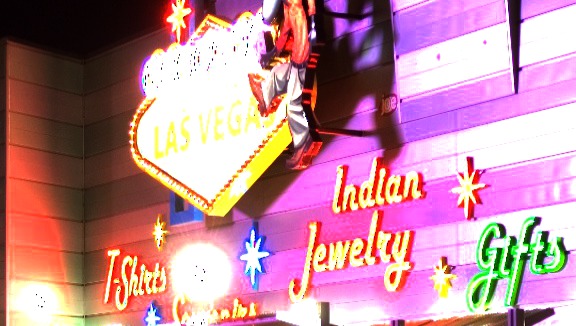}
    \end{subfigure}
    \begin{subfigure}[t]{0.13\linewidth}
        \centering
        \includegraphics[width=1.0\linewidth]{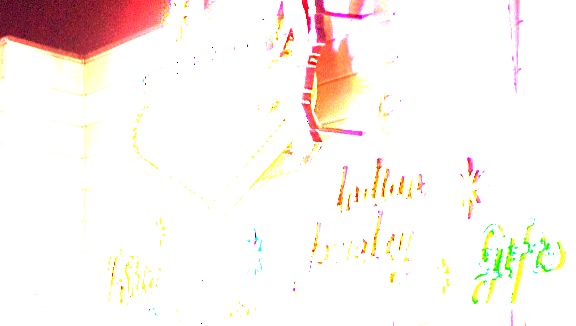}
    \end{subfigure}

    \begin{subfigure}[t]{0.13\linewidth}
        \centering
        \includegraphics[width=1.0\linewidth]{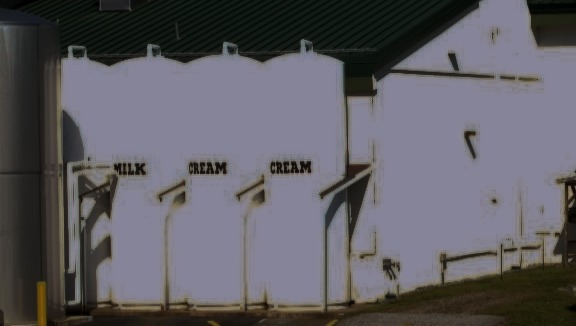}
    \end{subfigure}
    \begin{subfigure}[t]{0.13\linewidth}
        \centering
        \includegraphics[width=1.0\linewidth]{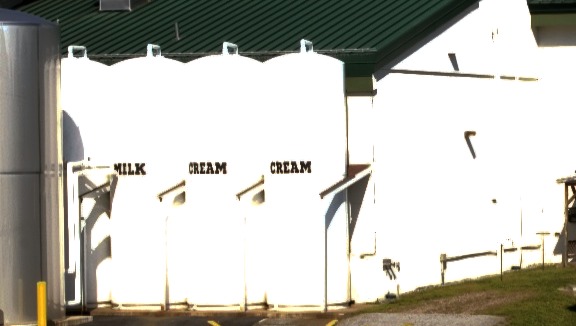}
        \caption{Ben Jerrys}\label{fig:exposures:benjerrys}
    \end{subfigure}
    \begin{subfigure}[t]{0.13\linewidth}
        \centering
        \includegraphics[width=1.0\linewidth]{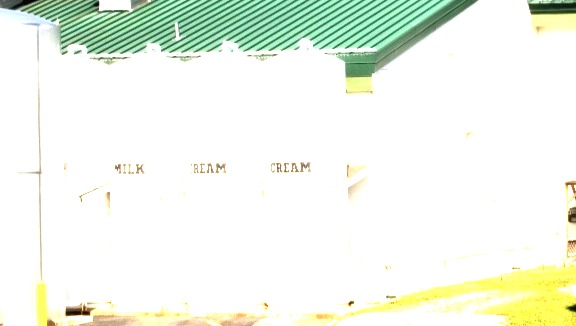}
    \end{subfigure}
    \begin{subfigure}[t]{0.13\linewidth}
        \centering
        \includegraphics[width=1.0\linewidth]{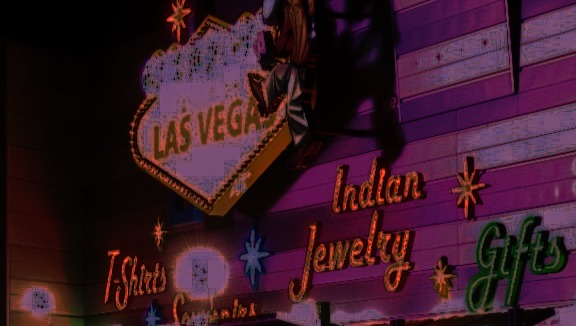}
    \end{subfigure}
    \begin{subfigure}[t]{0.13\linewidth}
        \centering
        \includegraphics[width=1.0\linewidth]{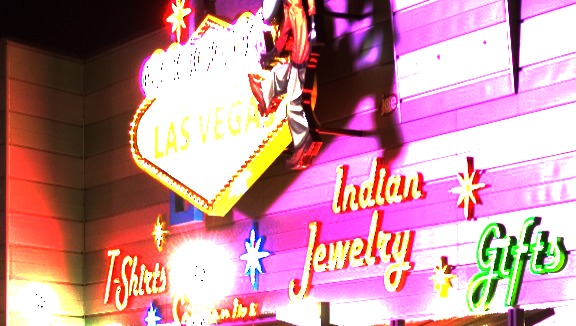}
        \caption{Las Vegas Store}\label{fig:exposures:lasvegas}
    \end{subfigure}
    \begin{subfigure}[t]{0.13\linewidth}
        \centering
        \includegraphics[width=1.0\linewidth]{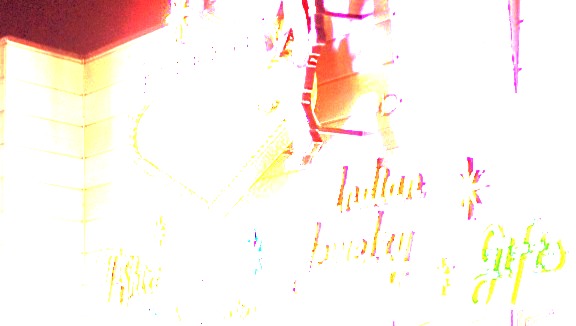}
    \end{subfigure}

    \begin{subfigure}[t]{0.40\linewidth}
        \centering
        \includegraphics[width=0.325\linewidth]{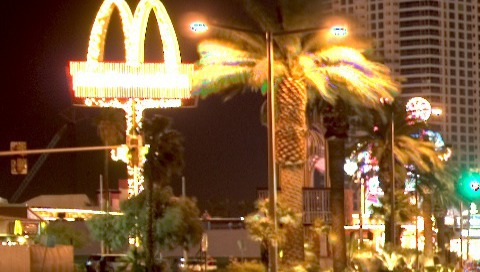}
    \end{subfigure}
    \begin{subfigure}[t]{0.40\linewidth}
        \centering
        \includegraphics[width=0.325\linewidth]{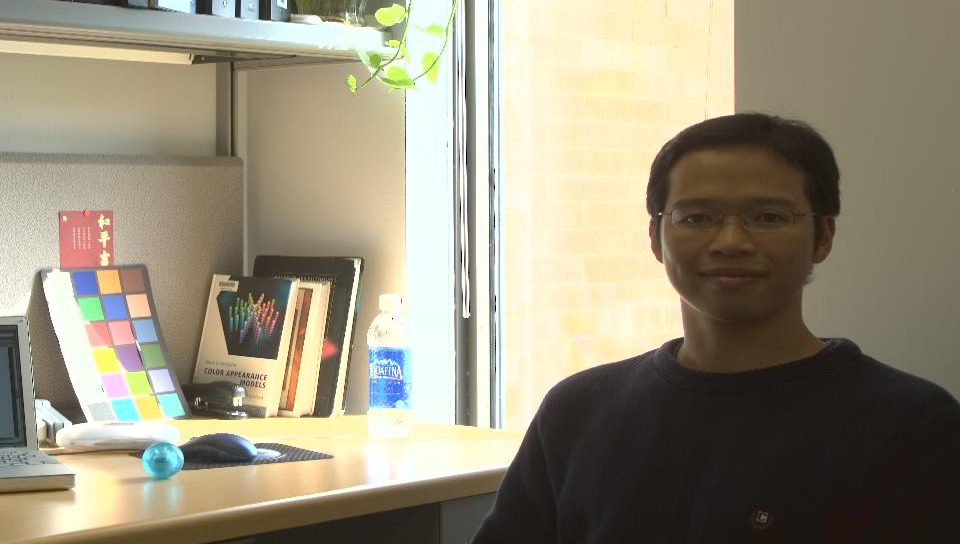}
    \end{subfigure}

    \begin{subfigure}[t]{0.13\linewidth}
        \centering
        \includegraphics[width=1.0\linewidth]{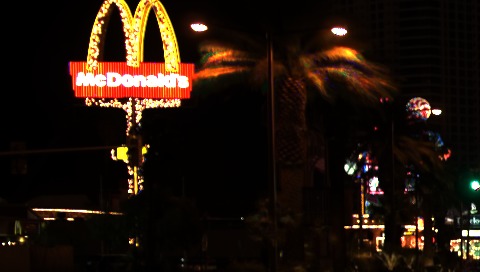}
    \end{subfigure}
    \begin{subfigure}[t]{0.13\linewidth}
        \centering
        \includegraphics[width=1.0\linewidth]{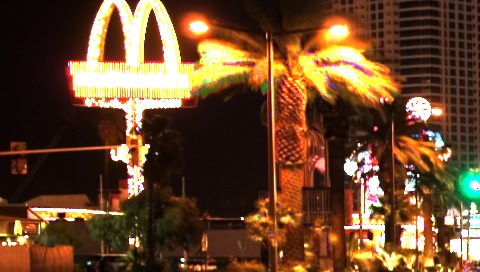}
    \end{subfigure}
    \begin{subfigure}[t]{0.13\linewidth}
        \centering
        \includegraphics[width=1.0\linewidth]{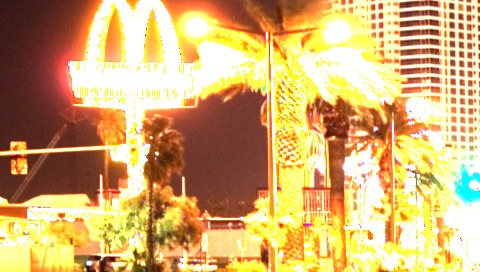}
    \end{subfigure}
    \begin{subfigure}[t]{0.13\linewidth}
        \centering
        \includegraphics[width=1.0\linewidth]{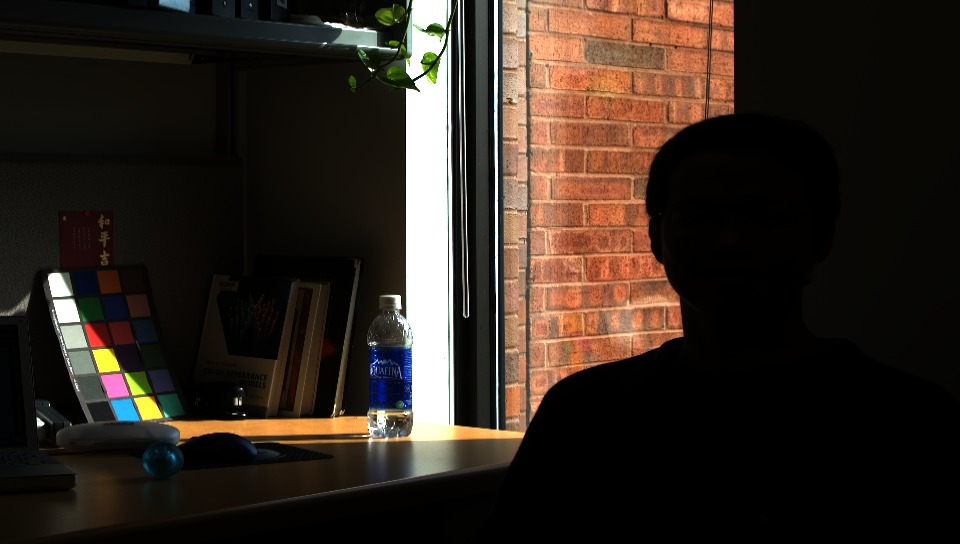}
    \end{subfigure}
    \begin{subfigure}[t]{0.13\linewidth}
        \centering
        \includegraphics[width=1.0\linewidth]{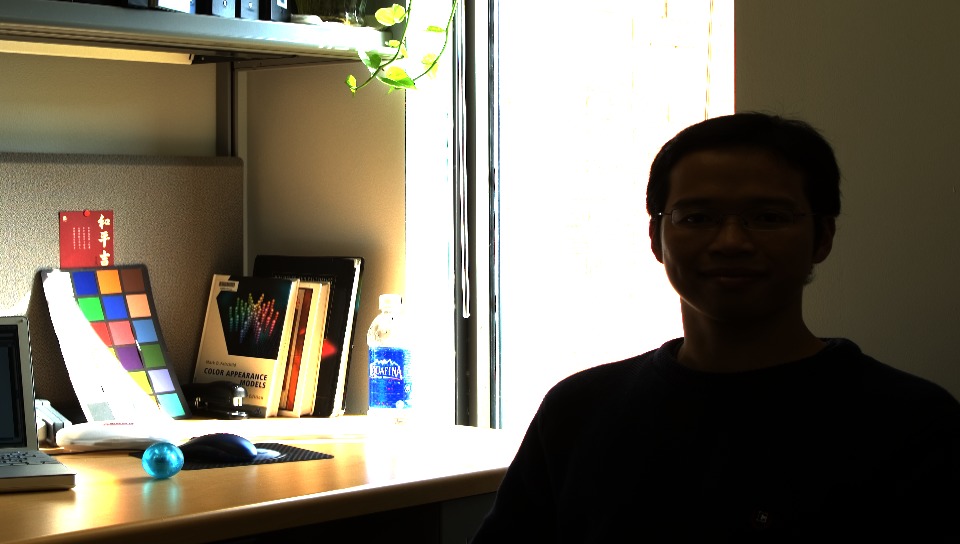}
    \end{subfigure}
    \begin{subfigure}[t]{0.13\linewidth}
        \centering
        \includegraphics[width=1.0\linewidth]{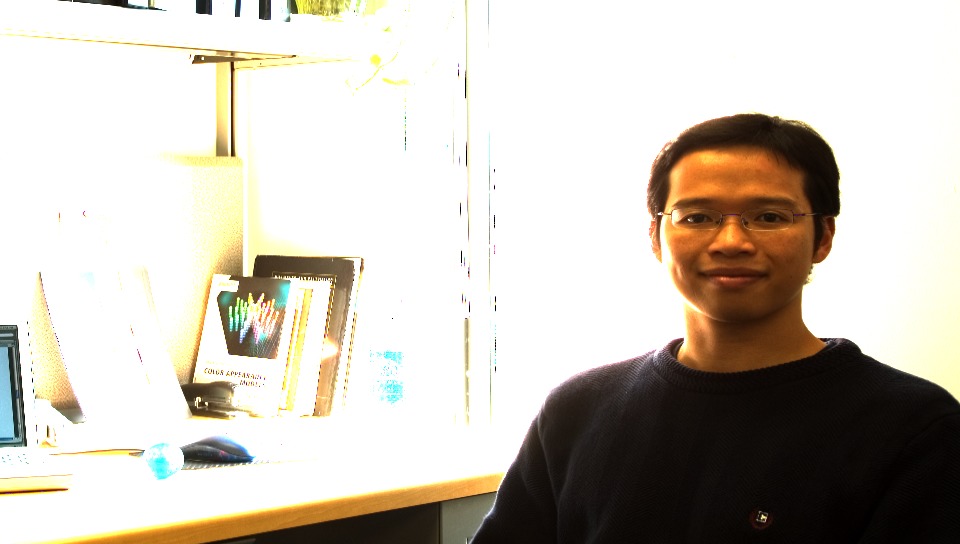}
    \end{subfigure}

    \begin{subfigure}[t]{0.13\linewidth}
        \centering
        \includegraphics[width=1.0\linewidth]{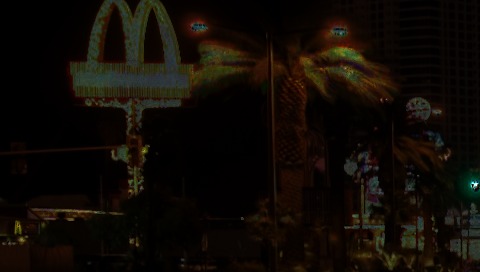}
    \end{subfigure}
    \begin{subfigure}[t]{0.13\linewidth}
        \centering
        \includegraphics[width=1.0\linewidth]{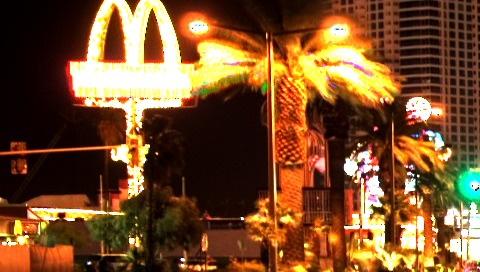}
    \end{subfigure}
    \begin{subfigure}[t]{0.13\linewidth}
        \centering
        \includegraphics[width=1.0\linewidth]{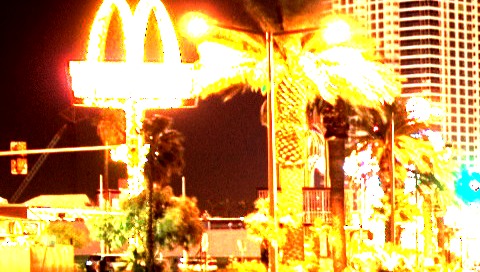}
    \end{subfigure}
    \begin{subfigure}[t]{0.13\linewidth}
        \centering
        \includegraphics[width=1.0\linewidth]{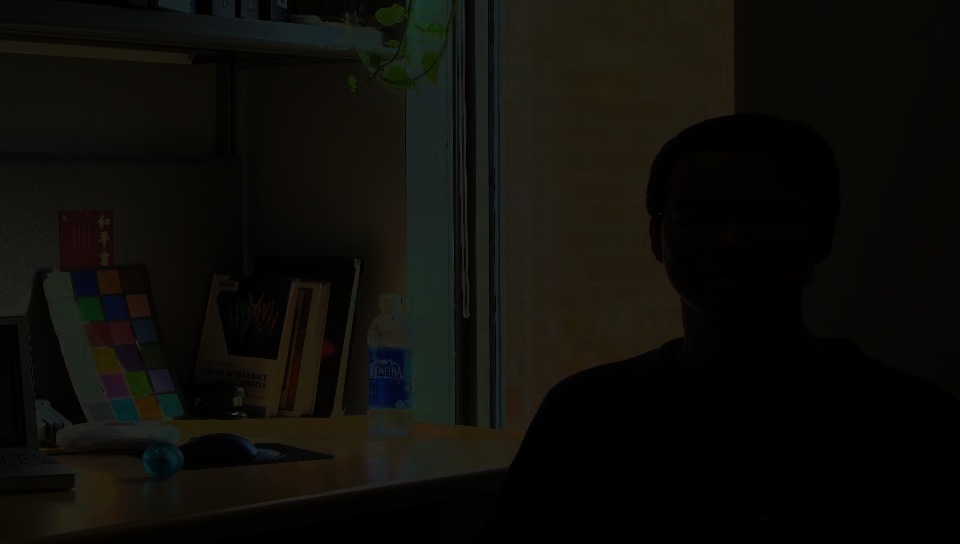}
    \end{subfigure}
    \begin{subfigure}[t]{0.13\linewidth}
        \centering
        \includegraphics[width=1.0\linewidth]{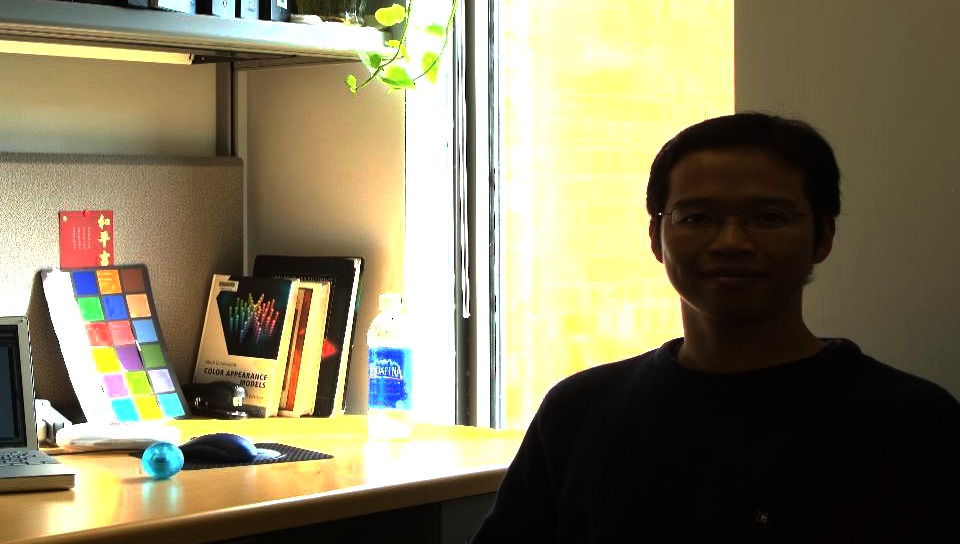}
    \end{subfigure}
    \begin{subfigure}[t]{0.13\linewidth}
        \centering
        \includegraphics[width=1.0\linewidth]{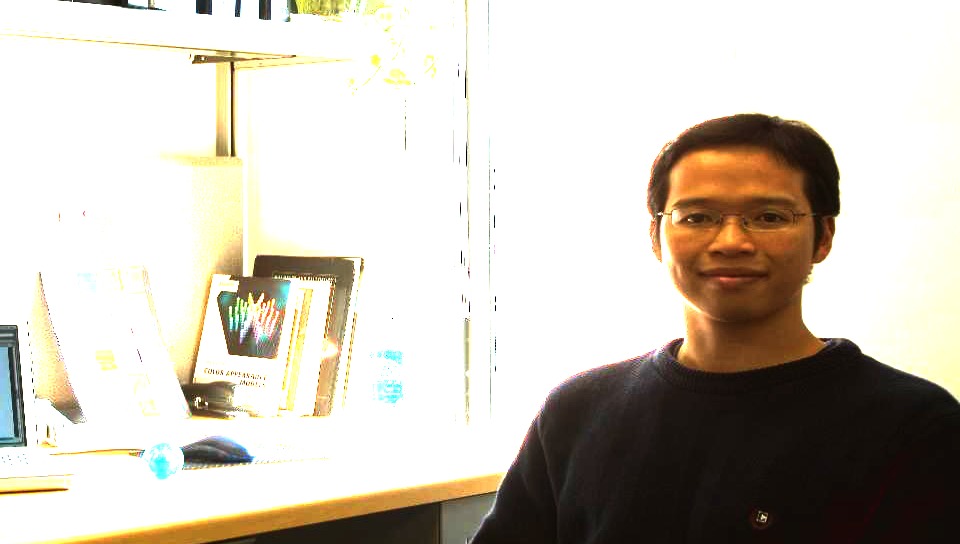}
    \end{subfigure}

    \begin{subfigure}[t]{0.13\linewidth}
        \centering
        \includegraphics[width=1.0\linewidth]{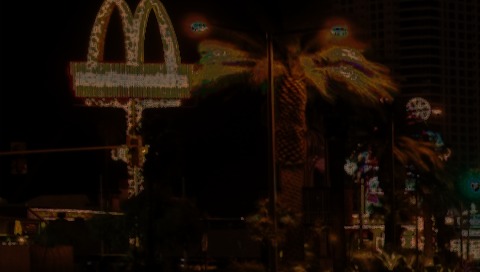}
    \end{subfigure}
    \begin{subfigure}[t]{0.13\linewidth}
        \centering
        \includegraphics[width=1.0\linewidth]{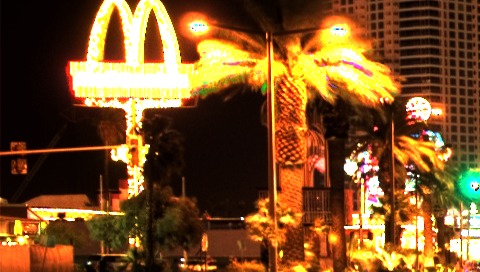}
        \caption{Peppermill}\label{fig:exposures:peppermill}
    \end{subfigure}
    \begin{subfigure}[t]{0.13\linewidth}
        \centering
        \includegraphics[width=1.0\linewidth]{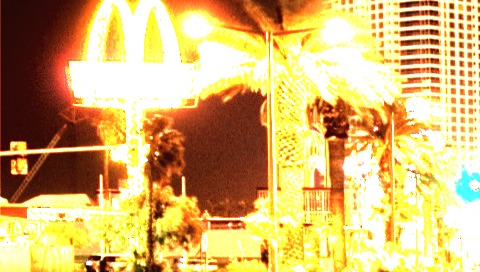}
    \end{subfigure}
    \begin{subfigure}[t]{0.13\linewidth}
        \centering
        \includegraphics[width=1.0\linewidth]{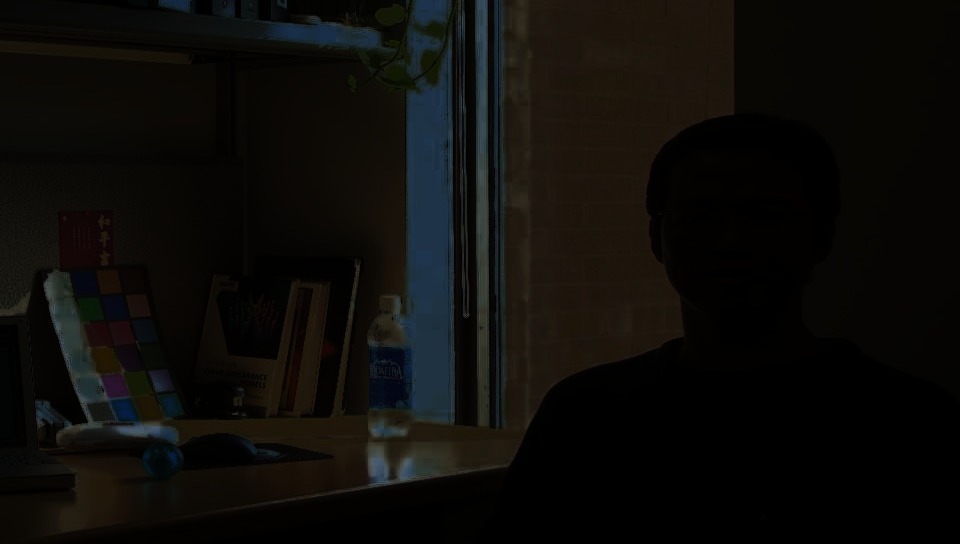}
    \end{subfigure}
    \begin{subfigure}[t]{0.13\linewidth}
        \centering
        \includegraphics[width=1.0\linewidth]{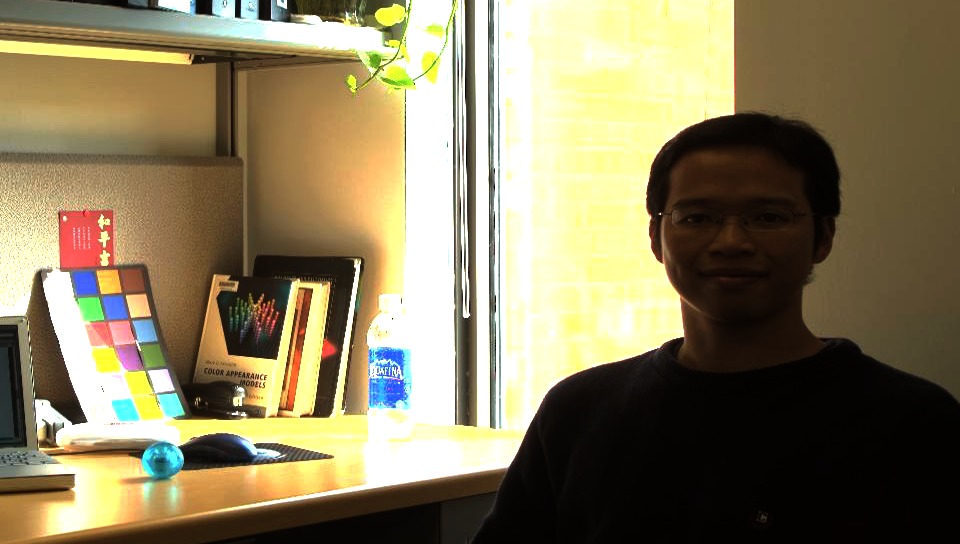}
        \caption{Willy Desk}\label{fig:exposures:willydesk}
    \end{subfigure}
    \begin{subfigure}[t]{0.13\linewidth}
        \centering
        \includegraphics[width=1.0\linewidth]{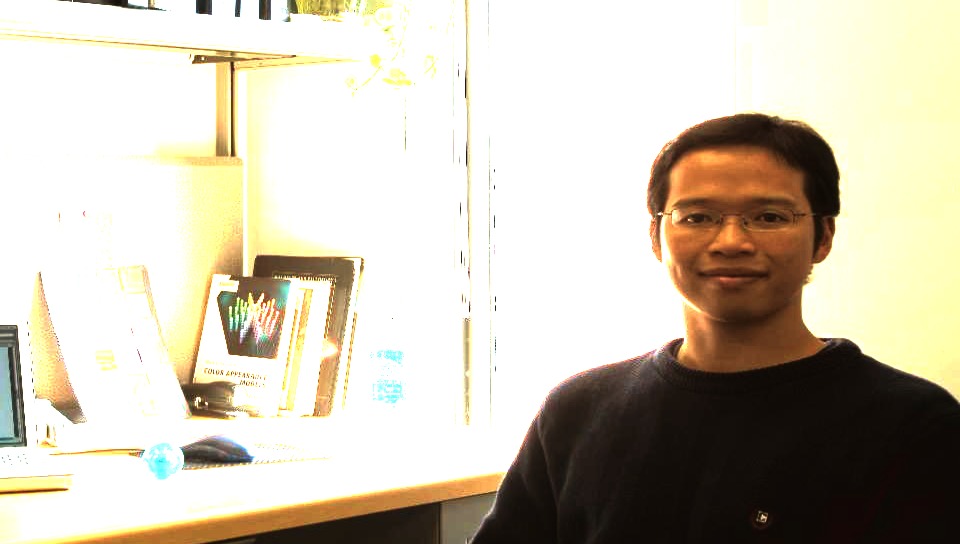}
    \end{subfigure}
    \caption{\tc{Examples of expanded images using EXP and EIL at three
    different exposures. The examples are cropped from larger images, showing
    under various lighting conditions and from different scenes. The top row of
    each sub-figure shows the input LDR created with \textit{culling}. The
    second row of each sub-figure shows the exposures of the original HDR. The
    following row shows exposures of predicted HDR using EIL. The last row
    shows exposures of predicted HDR using EXP.}}\label{fig:exposures}
\end{figure*}

Figure~\ref{fig:hallucinations_low} and Figure~\ref{fig:hallucinations_grotto}
show single exposure slices (both these cases are from low exposure slices) from the predicted HDRs for the four CNN
architectures. The input LDRs were created with \textit{culling} and are shown in the respective sub figure (f). It is clear
that UNT and COL have issues with blocking or banding and information bleeding, and this
can be observed, to a certain extent, for EIL as well, but to a much lesser degree.
\tc{Figure~\ref{fig:exposures} presents predictions at multiple exposures
comparing EXP and EIL. The images contain saturated areas of different
sizes as well as different combinations of saturated channels.
Figure~\ref{fig:exposures:benjerrys} contains blue pixels which after exposure
(scaling and clipping at 255) only have their B channel saturated (e.g.\ a pixel
[x, x, 243] becomes [x+y, x+y, 255] where B is clipped at 255).
Figure~\ref{fig:exposures:lasvegas} contains saturated purple pixels, where
both the R and B channels are clipped. Figure~\ref{fig:exposures:willydesk}
contains a saturated colour chart. It can be noticed that EXP tries to
minimize the bleeding of information into large overexposed areas, recovering
high frequency contrast, for example around text. It is also worth noting that
artefacts around sharp edges are not completely eliminated, but are much less
pronounced and with a much smaller extend.}

\begin{figure}[htb]
    \centering
    \includegraphics[width=1.0\linewidth]{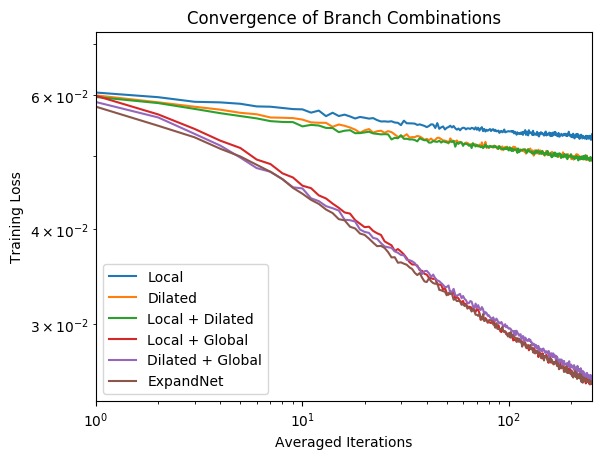}
    \caption{\tc{Training convergence for all the possible combinations of
    branches. Each point is an average of 10,000 gradient steps for a total of
    254,000 steps, the equivalent of 10,000 epochs (each epoch has 254
    mini-batches). Axes are logarithmic.}}\label{fig:branch_conv}
\end{figure}

\subsection{\textbf{Further Investigation}}
\label{sec:results:further}
\noindent
\tc{\textbf{Data Augmentation}: The method used to generate input-output pairs
significantly affects the end result. To demonstrate, the ExpandNet architecture
was trained on LDR inputs generated using only the \textit{Photoreceptor} TMO
(EXP-Photo). In this case it consistently underperforms when tested against EXP trained with
all the TMOs mentioned in Section~\ref{sec:dataset}, giving an average PSNR of
$19.93$ for display-referred \textit{culling}. However, if the testing is done
on LDR images produced not by \textit{culling}, but instead
\textit{Photoreceptor}, then EXP-Photo produces significantly better results
(PSNR of $24.28$ vs $21.52$ for EXP) since it was specialized to invert
the \textit{Photoreceptor} TMO. This can be useful if, for example, to convert
images captured by commercial mobile phones which are stored as tone mapped images using
a particular tone mapper back to HDR.}

\tc{To further investigate the effects of data augmentation, a network was
trained using Camera Response Functions (CRFs) in addition to the TMOs 
used for EXP reported in the previous section. Following the Deep Reverse Tone Mapping~\cite{endo2017drtmo}, the same
database of CRFs was used~\cite{grossberg2003crf}, and the same method of obtaining five representative
CRFs by k-means clustering was adopted.
The results do not show any improvement and are almost identical
to EXP on all metrics (within $1\%$). This might be because CRFs are monotonically increasing
functions, which can be approximated in many cases by the randomized
exposure and gamma TMO used in the initial set of results.}

\noindent \tc{\textbf{Branches}: To gain insight on the effect of the individual
branches and further motivate the three-branch architecture, different branch combinations were trained from scratch.
Figure~\ref{fig:branch_conv} shows the training convergence for 10,000 epochs.
It is evident that the global branch which is fused with each pixel makes the
largest contribution. On average, the full ExpandNet architecture is the
quickest to converge and has the lowest loss. The combination of the local and
dilation branches improves the performance of each one individually.}

\tc{We can further understand the architecture by comparing
figures~\ref{fig:branches} and~\ref{fig:branch_conv}. The performance of
Dilated + Global is comparable to that of Local + Global, even though
figure~\ref{fig:branches:LG} is visually much better than
\ref{fig:branches:DG}. This is because the images from
figure~\ref{fig:branches} are predictions from an ExpandNet with all branches
(some zeroed out when predicting), where the local and dilated branches have
acquired separate scales of focus during training (high and medium frequencies
respectively). In figure~\ref{fig:branch_conv}, where each one is trained
individually, these scales are not separated; each branch tries to learn all
the scales simultaneously. Separating scales in the architecture leads to
improved performance.}

\section{\textbf{Conclusions}}

This paper has introduced a method of expanding single exposure LDR content to
HDR via the use of CNNs. The novel three branch architecture provides a
dedicated solution for this type of problem as each of the branches account for
different aspects of the expansion. Via a number of metrics it was shown that
ExpandNet mostly outperforms the traditional expansion operators. Furthermore,
it performs better than non-dedicated CNN architectures based on UNT and
COL. \tc{Compared to other dedicated CNN methods
\cite{eilertsen2017cnn,endo2017drtmo} it does well in certain cases, exhibiting
fewer artefacts, particularly for content which is heavily under and over
exposed.} On the whole, ExpadNet is complementary to EIL which is designed to
expand the saturated areas and does very well in such cases. Furthermore, EIL
has a smaller memory footprint. ExpandNet has shown that a dedicated
architecture can be employed without the need of upsampling to convert HDR to
LDR, however, further challenges remain. \tc{To completely remove artefacts
further investigation is required, for example in the receptive fields of the
networks.} Dynamic methods may require further careful design to maintain
temporal coherence and Long Short Term Memory
networks~\cite{hochreiter1997lstm} might provide the solution for such content.

\section*{Acknowledgements}

Debattista is partially supported by a Royal Society Industrial Fellowship (IF130053).
Marnerides is funded by the EPSRC.

\section*{Revisions}

\textbf{v2}: Results in figure~\ref{fig:boxplots_scene_culling} and LAN (culling)
in Table~\ref{table:resultsscene} was corrected. The changes are minor and
do not alter the outcomes and conclusions.

\bibliographystyle{eg-alpha-doi}

\bibliography{expandnet}

\end{document}